\documentclass{ieeeaccess}
\usepackage{cite}
\usepackage{amsmath,amssymb,amsfonts}
\usepackage{graphicx}
\usepackage{textcomp}
\usepackage{balance}
\usepackage{booktabs} 
\usepackage{multirow} 
\usepackage{arydshln}   
\usepackage[table,dvipsnames]{xcolor} 
\usepackage[ruled,vlined]{algorithm2e}
\usepackage{bm}

\usepackage[colorlinks=true,linkcolor=blue,citecolor=blue]{hyperref}

\makeatletter
\AtBeginDocument{\DeclareMathVersion{bold}
\SetSymbolFont{operators}{bold}{T1}{times}{b}{n}
\SetSymbolFont{NewLetters}{bold}{T1}{times}{b}{it}
\SetMathAlphabet{\mathrm}{bold}{T1}{times}{b}{n}
\SetMathAlphabet{\mathit}{bold}{T1}{times}{b}{it}
\SetMathAlphabet{\mathbf}{bold}{T1}{times}{b}{n}
\SetMathAlphabet{\mathtt}{bold}{OT1}{pcr}{b}{n}
\SetSymbolFont{symbols}{bold}{OMS}{cmsy}{b}{n}
\renewcommand\boldmath{\@nomath\boldmath\mathversion{bold}}}
\newcommand{\coolname}{\text{Point-PNG}}
\newcommand{\transnet}{\text{COPE}}

\newcommand{\ie}{\emph{i.e.},}

\newsavebox{\capx}\sbox{\capx}{$\scriptstyle\mathbf{x}_i$}
\newsavebox{\capxplus}\sbox{\capxplus}{$\scriptstyle\mathbf{x}_i^{+}$}
\newsavebox{\capz}\sbox{\capz}{$\scriptstyle\mathbf{z}_i$}
\newsavebox{\capzplus}\sbox{\capzplus}{$\scriptstyle\mathbf{z}_i^{+}$}
\newsavebox{\capThetag}\sbox{\capThetag}{$\scriptstyle\Theta_g$}
\newsavebox{\capThetagr}\sbox{\capThetagr}{$\scriptstyle\Theta_{g_r}$}
\newsavebox{\capg}\sbox{\capg}{$\scriptstyle{}g$}
\newsavebox{\capgr}\sbox{\capgr}{$\scriptstyle{}g_r$}
\newsavebox{\capneg}\sbox{\capneg}{$\scriptstyle\tilde{\mathbf{z}}_i^{\,g_r}$}

\newsavebox{\capLoss}\sbox{\capLoss}{$\scriptstyle\mathcal{L}_{\coolname{}}$}
\newsavebox{\capAnchor}\sbox{\capAnchor}{$\scriptstyle\frac{\Theta_g\mathbf{z}_i}{\|\Theta_g\mathbf{z}_i\|}$}
\newsavebox{\capPseudo}\sbox{\capPseudo}{$\scriptstyle\frac{\Theta_{g_r} \mathbf{z}_i}{\|\Theta_{g_r} \mathbf{z}_i\|}$}
\def\cred{\textcolor{black}}
\def\cblue{\textcolor{black}}
\colorlet{forestgreen}{ForestGreen} 
\makeatother

\def\BibTeX{{\rm B\kern-.05em{\sc i\kern-.025em b}\kern-.08em
    T\kern-.1667em\lower.7ex\hbox{E}\kern-.125emX}}

\begin{document}
\history{Date of publication xxxx 00, 0000, date of current version xxxx 00, 0000.}
\doi{10.1109/ACCESS.2024.0429000}

\title{\coolname: Conditional Pseudo‑Negatives Generation for Point Cloud Pre-Training }

\author{
\uppercase{Sutharsan Mahendren}\authorrefmark{1,2},
\uppercase{Saimunur Rahman}\authorrefmark{1},
\uppercase{Piotr Koniusz}\authorrefmark{1,3},\\
\uppercase{Tharindu Fernando}\authorrefmark{2},
\uppercase{Sridha Sridharan}\authorrefmark{2},
\uppercase{Clinton Fookes}\authorrefmark{2},
\uppercase{Peyman Moghadam}\authorrefmark{1,2}
}

\address[1]{CSIRO Robotics, DATA61, CSIRO, Australia. (e-mail: \{firstname.lastname\}@csiro.au)}
\address[2]{School of Electrical Engineering and Robotics, Queensland University of Technology (QUT), Brisbane, Australia.}
\address[3]{Australian National University.}

\markboth
{Mahendren \headeretal: \coolname{}: Conditional Pseudo‑Negatives Generation for Point Cloud Pre-Training }
{Mahendren \headeretal: \coolname{}: Conditional Pseudo‑Negatives Generation for Point Cloud Pre-Training }

\corresp{Corresponding author:  Sutharsan Mahendren (e-mail: sutharsan.mahendren@csiro.au).}

\begin{abstract}
We propose \coolname{},  a novel self-supervised learning framework that generates conditional pseudo‑negatives in the latent space to learn point cloud representations that are both discriminative and transformation‑sensitive.
Conventional self-supervised learning methods focus on achieving invariance, discarding transformation-specific information. 
Recent approaches incorporate transformation sensitivity by explicitly modeling relationships between original and transformed inputs. However, they often suffer from an invariant-collapse phenomenon, where the predictor degenerates into identity mappings, resulting in latent representations with limited variation across transformations. To address this, we propose \coolname{} that explicitly penalizes invariant collapse through pseudo-negatives generation, enabling the network to capture richer transformation cues while preserving discriminative representations.
To this end, we introduce a parametric network, \underline{CO}nditional \underline{P}seudo-Negatives \underline{E}mbedding (\transnet{}), which learns localized displacements induced by transformations within the latent space. A key challenge arises when jointly training \transnet{} with the MAE, as it tends to converge to trivial identity mappings. To overcome this, we design a loss function based on pseudo-negatives conditioned on the transformation, which penalizes such trivial invariant solutions and enforces meaningful representation learning.
We validate \coolname{} on shape classification and relative pose estimation tasks, showing competitive performance on ModelNet40 and ScanObjectNN under challenging evaluation protocols, and achieving superior accuracy in relative pose estimation compared to supervised baselines.

\end{abstract}

\begin{keywords}
Representation learning, Point Clouds, contrastive learning, Self-Supervised Learning
\end{keywords}

\titlepgskip=-21pt

\maketitle
\section{Introduction}
\label{sec:intro}
Self-Supervised Learning (SSL) has emerged as a crucial approach in point cloud analysis, allowing models to learn robust and generalizable representations from large volumes of unlabeled 3D data. With the growing accessibility of affordable 3D sensors and scanning devices, point clouds have become a widely used data format in applications such as autonomous driving, robotics, and augmented reality.
SSL methods have achieved remarkable success in various downstream tasks involving point clouds, including 3D object classification \cite{afham2022crosspoint, abouzeid2023point2vec, pang2022masked, zhang2022point, sun20233d}, 3D scene understanding \cite{zhan2019self, xie2020pointcontrast, cheng2025lsv}, and 3D part segmentation \cite{afham2022crosspoint, abouzeid2023point2vec, pang2022masked, zhang2022point}. 
Traditional SSL methods often focus on learning invariant representations to input perturbations or transformations. 
However, enforcing strict invariance to transformations may suppress critical geometric cues, such as orientation or spatial displacement, that are essential for tasks in robotics and other autonomous applications where the structure and dynamics of the physical world must be preserved.

Recent approaches such as EquiMod~\cite{devillers2023equimod} and SIE~\cite{garrido2023sie} address this limitation by introducing sensitivity to transformations (\ie{} equivariant objective), allowing models to explicitly encode the relationships between original and transformed point clouds in the latent space.
\cblue{ %
SIE~\cite{garrido2023sie} shows that the linear predictor used by EquiMod~\cite{devillers2023equimod} can suppress transformation channels, collapse into an identity mapping, and produce representations that are no more informative than those of purely invariant methods. To mitigate this shortcut, SIE replaces the linear layer with a bias-free hypernetwork whose weights are conditioned on the transformation. 
Although the hypernetwork conditions its weights on the transformation, it cannot guarantee that the predicted transformation deviates meaningfully from the identity. This limitation constrains the model’s ability to capture significant local displacements in the latent space. Consequently, the resulting representations may exhibit only limited variation in response to different transformations, reducing their sensitivity. This underscores the need for strong regularization mechanisms that actively promote responsiveness to transformations.
}

In this work, we propose a novel self-supervised learning framework, \coolname{}, which learns transformation-sensitive representations while preserving discriminative capacity. 
Our key contribution is a loss function that bootstraps pseudo-negatives conditioned on randomly sampled transformations, avoiding the risk of collapse to invariant solutions. 
These pseudo-negatives serve as reference embeddings generated using the predictor, capturing the localized variations in the embedding space caused by different transformations. By incorporating them into the training objective, we encourage the model to produce representations that are sensitive to input transformations while avoiding degenerate solutions.
To achieve this, we introduce the \underline{CO}nditional \underline{P}seudo Negatives \underline{E}mbedding network (\transnet{}). \transnet{} takes transformation parameters as input and outputs a weight used to linearly project the original embeddings. This conditional mechanism enables the generation of transformation specific pseudo negatives. We use these pseudo-negatives to regularize \transnet{} through our proposed loss function, ensuring that distinct transformations yield sufficiently different embeddings in the latent space.

An advantage of our method is that it formulates relative pose estimation as an optimization problem in the embedding space, using transformation-sensitive features learned via the \transnet{}. We introduce a novel inference algorithm that iteratively aligns source and target point cloud embeddings, achieving precise alignment even with large initial misalignments.

We validate the effectiveness of \coolname{} through comprehensive experiments on shape classification and relative pose estimation tasks. On the ModelNet40 and ScanObjectNN datasets, our method achieves state-of-the-art performance, particularly under challenging protocols. For relative pose estimation tasks, our framework excels in accurately estimating relative poses between point clouds, leading to superior performance compared to supervised baselines. 
For completeness, we also include a head-to-head comparison on a 3D-relevant image benchmark in which inputs are 2D renderings of 3D objects undergoing controlled 3D rotations. Although rendered in 2D, this benchmark evaluates sensitivity to 3D transformations. We compare against transformation-sensitive contrastive losses~\cite{devillers2023equimod,garrido2023sie} under the same backbone and protocol, evaluating the discriminativeness and rotation sensitivity of the learned features.

\section{Related Works}

\subsection{Self-Supervised Learning for Point Cloud}
Self-Supervised Learning (SSL) has become increasingly prevalent for leveraging unlabeled data, especially in the challenging domain of 3D point clouds. The inherent disorder and lack of predefined structure in point clouds necessitate diverse strategies for extracting meaningful representations. Among these, contrastive methods \cite{xie2020pointcontrast,zhang2021self,afham2022crosspoint, sanghi2020info3d} exploit correspondences across different views of point clouds to establish unsupervised pretraining frameworks using InfoNCE loss \cite{oord2018representation,chen20j}. In parallel, recent advancements have introduced transformer-based masked autoencoder approaches, such as PointBERT \cite{yu2022point}, which decodes discrete tokens from point patches, PointMAE \cite{pang2022masked} along with PointM2AE \cite{zhang2022point}, which directly predict the masked point patches rather than tokens and PointGPT~\cite{chen2023pointgpt}, which follows auto-regressive prediction of spatially ordered point patches. PCPN~\cite{yamada2024masked} adopts PointMAE \cite{pang2022masked} for pretraining on procedurally generated point clouds. ExpPoint-MAE~\cite{10497601}  compares masked autoencoding and momentum-contrast~\cite{9157636} pretraining for point cloud transformers, and introduces a strategic unfreezing fine-tuning schedule. 
Concurrently, LiDAR-specific SSL methods~\cite{krispel2024maeli,abdelsamad2025multi} adapt masked reconstruction objective to sensing characteristics.
Extending this paradigm, Point2Vec \cite{abouzeid2023point2vec} employs the principles of Data2vec \cite{baevski2022data2vec} to facilitate feature space reconstruction in point clouds. 
Recent works extend reconstruction across modalities, either rendering 3D into 2D with differentiable renderers~\cite{Zhu2025TPAMIPonderV2,yang2024unipad,wang2025unipre3d} or distilling multi-view image features~\cite{zhang2023learning} instead our work focus on single-modality representation learning.
Diverging from SSL techniques that mainly induce biases towards invariant feature learning or local relationships for reconstruction, our approach focuses on learning both invariant and equivariant representations, significantly enhancing the performance across both downstream tasks. 

\subsection{Supervised Equivariant Networks}

EPN~\cite{chen2021equivariant} and E$^2$PN~\cite{zhu2023e2pn} introduce SE(3)-equivariant architectures designed for point cloud analysis under supervised learning. EPN builds on KPConv~\cite{thomas2019kpconv} by incorporating group convolutions over discretized SO(3) to achieve rotation equivariance, while E$^2$PN~\cite{zhu2023e2pn} improves efficiency by using quotient representations on $S^2 \times \mathbb{R}^3$ with a permutation layer to recover full rotational structure. These models show strong performance on classification and relative pose estimation tasks but rely on labeled data. Additionally, Vector Neurons~\cite{deng2021vector} represent another notable class of SO(3)-equivariant networks, using vector-valued features that transform linearly under rotation. EquivReg~\cite{zhu2022correspondence} leverages Vector Neurons for correspondence-free point cloud registration, solving for relative rotations in closed-form within the equivariant feature space. Our method, by contrast, leverages equivariance as a self-supervisory signal for representation learning with a non-equivariant backbone. This enables training without labeled data while still retaining the benefits of transformation-aware learning. We compare against supervised baselines to demonstrate the advantages of our self-supervised framework.

\subsection{Sensitivity to Transformation in SSL methods}
Contrastive learning ~\cite{chen20j, 9157636, knights2023geoadapt} has significantly advanced self-supervised representation learning. However, enforcing strict invariance to data augmentations can limit the expressive power of the learned features, especially in robotic tasks where details about specific transformations are crucial. 
To overcome this limitation, image-based studies~\cite{dangovski2022equivariant, lee2021improving, devillers2023equimod, garrido2023sie} have explored enhancing models by incorporating sensitivity to transformations. 
One approach involves auxiliary tasks that require models to predict the transformations applied to the input data~\cite{gidaris2018unsupervised, lee2021improving, dangovski2022equivariant, scherr2022self}, which helps preserve transformation-specific information within the representations and leads to less invariant feature spaces. 
In LiDAR point clouds, PSA-SSL~\cite{nisar2025psa} addresses this by adding a self-supervised bounding box regression task in addition to contrastive objective to encode pose and size. 
However, these methods often lack guarantees of a consistent mapping between input transformations and changes in the latent representations~\cite{garrido2023sie}. 
Alternative approaches~\cite{gupta2024structuring, wang2024pose} integrate loss functions that balance invariance with sensitivity to augmentations, enabling models to remain responsive to transformations while maintaining robustness. 
Additionally, recent methods~\cite{devillers2023equimod, garrido2023sie,park2022sen} have developed models that directly map transformations in the input space to transformations in the latent space, employing external predictors to modify representations accordingly.
These methods~\cite{devillers2023equimod,garrido2023sie} identify the invariant-collapse problem, where predictors degenerate into identity mappings, effectively removing essential transformation cues from the learned representations. SIE~\cite{garrido2023sie} addresses this by replacing the linear predictor with a transformation-conditioned hypernetwork. However, without strong regularization, the predicted transformations of SIE can still converge toward the identity, leading to limited transformation sensitivity as shown by the proof in Appendix.
In contrast, we introduce a novel penalty loss based on transformation-conditioned pseudo-negatives that explicitly discourages collapse into invariant representations and promotes responsiveness to transformations. Since EquiMod~\cite{devillers2023equimod}, SEN~\cite{park2022sen}, and SIE~\cite{garrido2023sie} were originally developed for the image domain, we include additional experiments evaluating our method against these baselines on image-based tasks. The results confirm that our proposed penalty loss effectively mitigates invariant collapse and enhances transformation sensitivity.

\section{Proposed Approach }
\begin{figure*}[!th]
  \centering
\includegraphics[width=1.0\textwidth]{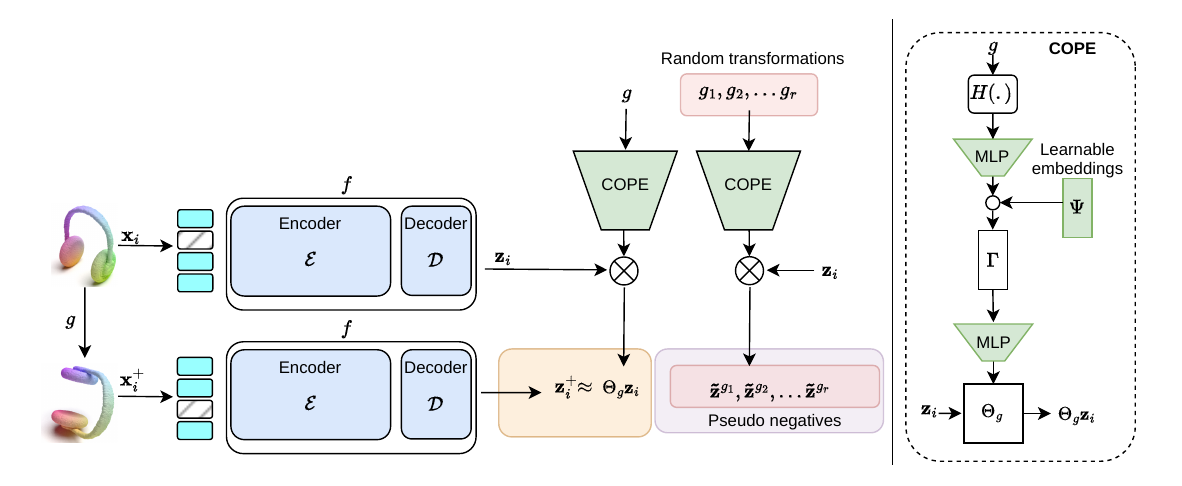}
    \caption{Overview of \coolname{}: It encodes a point cloud \usebox{\capx} and its transformed counterpart \usebox{\capxplus}, extracting their global representations \usebox{\capz} and \usebox{\capzplus}, respectively, using a shared transformer architecture. Simultaneously, \transnet{} network outputs the weight \usebox{\capThetag} and \{\usebox{\capThetagr}\} based on the input and random transformations \usebox{\capg} and \{\usebox{\capgr}\}. \{\usebox{\capThetagr}\} are used to generate pseudo-negatives \{\usebox{\capneg}\}.
    }
\label{fig.overall_method}
\end{figure*}

In this section, we introduce \coolname{}, a method designed to learn feature representations that are both discriminative and transformation-sensitive. Our framework builds on the Masked AutoEncoder (MAE) transformer, combining the strengths of predictive (MAE) and discriminative (contrastive) training paradigms. As illustrated in Fig.~\ref{fig.overall_method}, we model the relationship between input transformations and displacements in the embedding space using linear projections, controlled by a network \transnet{} that generates transformation-specific weights. The main network $f$ and \transnet{} are trained jointly. However, \transnet{} risks collapsing into the identity matrix, which would yield invariant representations. To prevent this, we introduce a novel loss function based on pseudo-negatives that regularizes \transnet{} and enforces equivariant projections during training.

Let $f$ be a parametric network that maps an input space $\mathcal{X}$ to the unit sphere. Here $f$ is a composition of an encoder $\mathcal{E}$ and a decoder $\mathcal{D}$, (\ie{} $f =  \mathcal{D} \circ \mathcal{E} $).
Let $\mathcal{G}$ be a distribution of possible transformations in the input space $ \mathcal{X}$, and $g$ denotes a transformation sampled from the distribution $\mathcal{G}$. 
For the given input point cloud $\mathbf{x}_i \in \mathcal{X}$, $g$ transforms it to a positive view $\mathbf{x}_i^+ = g(\mathbf{x}_i)$. The representations $\mathbf{z}$ of $\mathbf{x}_i, \mathbf{x}_i^+$ is defined as $\mathbf{z}_i = f(\mathbf{x}_i) \text{ and }  \mathbf{z}_i^+ = f(\mathbf{x}_i^+)$. 
Both $\mathbf{z}_i$ and $\mathbf{z}_i^+$ reside within the same representation space $\mathcal{Z}$ of $f$. 
We introduce an operator $u_g$ that models how the transformation $g$ affects the feature representation $\mathbf{z}_i$.

Contrastive learning methods, such as SimCLR \cite{chen20j}, typically aim to learn a representation that is invariant to perturbations or transformations of the input data. This is achieved by encouraging similar instances to have similar representations in the embedding space, while dissimilar instances are pushed apart.
This objective is equivalent to a combination of alignment $\mathcal{L}_{\text{align}}$ and uniformity $\mathcal{L}_{\text{unif}}$, as described in Eq. (\ref{eq:ssl}) below: 

\begin{equation}
\label{eq:ssl}
\begin{split}
    \mathcal{L}_{\text{align}} &= \frac{1}{N} \sum_{i=1}^{N} \left\|\mathbf{z}_i - \mathbf{z}_i^+\right\|_2^2; \\
    \mathcal{L}_{\text{unif}} &= \log \mathbb{E}_{(\mathbf{z}_i, \mathbf{z}_k)\sim \mathcal{Z}\times\mathcal{Z}} e^{-\left\|\mathbf{z}_i - \mathbf{z}_k\right\|_2^2 / \tau}; \\
    \mathcal{L}_{\text{ssl}} &= \mathcal{L}_{\text{align}} + \mathcal{L}_{\text{unif}}.
\end{split}
\end{equation}

Specifically, $\mathcal{L}_{\text{align}}$ promotes the invariance property, while $\mathcal{L}_{\text{unif}}$ regularizes the learning process by preventing dimensional collapse, where $f(x)$ becomes constant for all $x$~\cite{wang2020understanding}. Here $\tau$ is a temperature parameter.
While invariant SSL models facilitate the learning process by discarding the variations introduced by transformations, they often neglect potentially critical information that these variations might carry. To address this issue, a group of self-supervised learning methods \cite{devillers2023equimod, garrido2023sie, gupta2024structuring} has introduced the concept of \textit{sensitivity to transformation}, enabling the model to explicitly encode the relationships between the original and augmented data in the latent space.

These approaches introduce a modified alignment term, $\mathcal{L}_{\text{align}^*} = \frac{1}{N} \sum_{i=1}^{N} \left\|u_g(\mathbf{z}_i) - \mathbf{z}_i^{+}\right\|_2^2$, where $u_g$ defines the relationship between the feature representations $\mathbf{z}_i$ and their augmented counterparts $\mathbf{z}_i^{+}$ under the transformation $g$. When combined with the uniformity term from Eq. (\ref{eq:ssl}), this alignment term helps the model avoid dimensional collapse while enhancing its sensitivity to transformations.
A trivial solution to $\min_f \mathcal{L}_{\text{align}^*}$ is for all transformations $\forall g \in \mathcal{G}$ to collapse into the same embedding space (\ie{} $u_g(\mathbf{z}_i) = \mathbf{z}_i$), making the embeddings invariant to transformations. To avoid this, our method, \coolname{}, introduces a constraint that produces localized displacements in the embedding space, ensuring that: $u_{g_{r}}(\mathbf{z}_i) \neq u_{g_s} (\mathbf{z}_i) ;\quad  (g_r, g_s) \sim \mathcal{G}\times\mathcal{G} :   g_r \neq g_s .$
\\
\subsection{Sensitivity to Transformation through Pseudo-Negatives}
To effectively model the relationship $u_g$, we introduce \underline{CO}nditional \underline{P}seudo Negatives \underline{E}mbedding (\transnet{}) network, a parameter efficient row-wise weight generation module with shared parameters.
Specifically, \transnet{} takes transformation parameter $g$ as an input and outputs the weight $\Theta_g= \transnet(g) \in \mathbb{R}^{d \times d}$.
This weight is used as a linear projection on $\mathbf{z}_i$, which learns the localized displacements due to the transformation $g$. 

In our framework, \transnet{} is jointly optimized with the main network $f$. Here, \transnet{} learns to approximate the effect of transformations through its output weights $\Theta_g$, while $f$ must learn representations $\mathbf{z}_i$ and $\mathbf{z}_i^{+}$ that satisfy this linear constraint dependent on the transformations $g$. This is a challenging task because it requires $f$ and \transnet{} to closely cooperate. 

A potential shortcut for \transnet{} is to produce identity weights for all transformations, $\Theta_g = I$, making the features invariant to transformations. This trivial solution undermines our goal of learning transformation-sensitive representations. To avoid this, we introduce the process of \textit{conditional generation of pseudo-negatives} using \transnet{}. With pseudo-negatives, we regularize \transnet{} and prevent it from collapsing into the identity. This approach ensures that different transformations produce different projections in the embedding, maintaining the sensitivity to the transformations.
Specifically, we generate pseudo-negatives by applying \transnet{} to a set of randomly sampled transformations $\{g_r\}$ from $\mathcal{G}$, where $\{g_r\}$ are different from the input transformation $g$. For each $g_r$, the corresponding weight $\Theta{g_r}$ with the \transnet{} is computed and generated the pseudo-negative embedding $\mathbf{\tilde{{z}}}_{i}^{g_r}$:
\begin{equation} 
\label{eq: gen_pseudo_neg} 
\mathbf{\tilde{z}}_{i}^{g_r} = \frac{\transnet(g_r)\mathbf{z}_i}{\|\transnet(g_r)\mathbf{z}_i\|} = \frac{\Theta_{g_r} \mathbf{z}_i}{\|\Theta_{g_r} \mathbf{z}_i\|}.
\end{equation}

We incorporate these pseudo-negatives into our novel loss function $\mathcal{L}_{\text{cope}}$ to regularize \transnet{}. 
\begin{equation} 
\label{eq: l_cope}
\begin{split}
\mathcal{L}_{\text{cope}} &= \frac{1}{N} \sum_{i=1}^{N}\log \Biggl[\sum_{r=1}^{M} e^{-\left\| \frac{\Theta_g \mathbf{z}_i}{\|\Theta_g \mathbf{z}_i\|} - \mathbf{\tilde{{z}}}_{i}^{g_r}  \right\|_2^2 / \tau} + 
\\
&\qquad\qquad\quad
e^{-\left\| \frac{\Theta_g \mathbf{z}_i}{{\|\Theta_g \mathbf{z}_i\|}} - \mathbf{z}_i^{+} \right\|_2^2 / \tau} \Biggr],
\end{split}
\end{equation}
where $M$ is the number of pseudo-negatives. This loss penalizes the model for mapping different transformations to similar embeddings, thus preventing the trivial solution. 
Consider the case of perfect alignment where $\mathbf{z}_i^{+} = \frac{\Theta_g \mathbf{z}_i}{\|\Theta_g \mathbf{z}_i\|}$. Minimizing $\mathcal{L}_{\text{cope}}$ is equivalent to optimizing $\log \left[ \sum_{r=1}^{M} e^{-\left\| \frac{\Theta_g \mathbf{z}_i}{{\|\Theta_g \mathbf{z}_i\|}} - \mathbf{\tilde{{z}}}_{i}^{g_r}  \right\|_2^2 / \tau} + 1 \right]$,
which effectively maximizes the pairwise distances between $\{\Theta_{g_r} \mathbf{z}_i\}$. Achieving this requires diverse $\{\Theta_{g_r}\}$ from the \transnet{}, thereby penalizing trivial invariant solutions.
We define the final loss $\mathcal{L}_{\coolname}$ as follows: 
\begin{equation}
    \mathcal{L}_{\coolname} = \mathcal{L}_{\text{align}^*} +  \beta \mathcal{L}_{\text{cope}} + (1 - \beta) \mathcal{L}_{\text{unif}}.
\end{equation}

\begin{figure}[!t]
  \centering
  \hspace*{1.cm} 
  \includegraphics[width=0.9\linewidth]{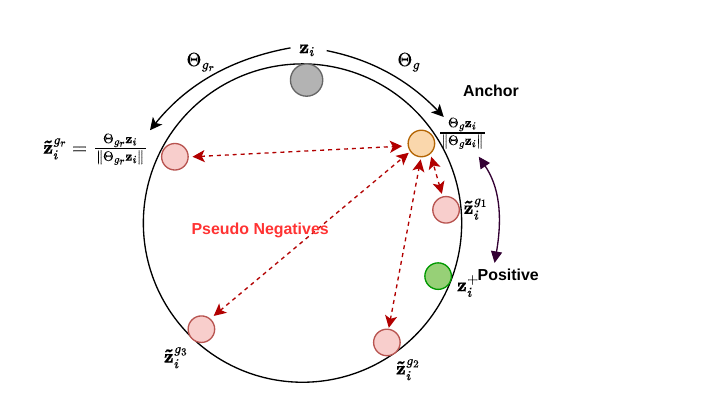}
  \caption{Conceptual visualization of the loss \usebox{\capLoss}. Using \transnet{}, we generate the anchor \usebox{\capAnchor} and a set of pseudo-negatives \usebox{\capPseudo} for the corresponding positive \usebox{\capzplus}. As per our main goal, due to these pseudo-negatives, the model remains sensitive to different transformations rather than becoming completely invariant to them.}
\label{fig: pseudo_negatives}
\vspace{-1em}
\end{figure}
Specifically, $\mathcal{L}_{\text{unif}}$ encourages discriminative features and $\mathcal{L}_{\text{cope}}$ preserves the sensitivity to transformations within the feature space. This combination enables the model to learn discriminative and sensitive representations to input transformations. 
Fig. \ref{fig: pseudo_negatives} illustrates the objective of our loss in terms of the anchor, the positive, and the pseudo-negatives. 
The loss $L_{\text{align}^*}$ encourages the positive $\mathbf{z}_i$ to be pulled towards the anchor.
$\mathcal{L}_{\text{cope}}$ encourages the pseudo-negatives $\mathbf{\tilde{{z}}}_{i}^{g_r}$ to be pushed away from the anchor $\frac{\Theta_g \mathbf{z}_i}{\|\Theta_g \mathbf{z}_i\|}$. For a given $\mathbf{z}_i$ all the pseudo-negatives $\{\frac{\Theta_{g_r} \mathbf{z}_i}{\|\Theta_{g_r} \mathbf{z}_i\|}\}$ only depend on the output of the \transnet{} $\Theta_{g_r}$ network. This avoids $\Theta_{g}$ collapsing into identity, preventing the undesired invariant solution. \cblue{Unlike SIE~\cite{garrido2023sie} whose equivariance is mainly enforced through an alignment loss, a term that trivially vanishes when the predictor collapses to the identity, $\mathcal{L}_{COPE}$ conditioned with the Pseudo-Negatives reaches its maximum when predictor collapse to identity. The full proof appears in the Appendix.}

\subsection{COPE Network Architecture}
\transnet{} generates linear weights conditioned on the transformation set ${\{g}\}$. In our work, we consider rotations as transformations and utilize quaternions to represent them as inputs to the \transnet{}. The network architecture is illustrated in the right section of Fig. \ref{fig.overall_method}. 
First, we embed $g$ into a harmonic space  $H(\cdot)$, using a set of high-frequency functions \cite{mildenhall2021nerf}. Subsequently, these embedded quaternions are processed through an MLP network to extract transformation-sensitive embeddings, denoted as $\mathbf{h}_g' \in \mathbb{R}^{\frac{d}{k}}$. Independently estimating each element of $\Theta_g$ is computationally expensive. Therefore, we utilize shared learnable embedding vectors $\Psi \in \mathbb{R}^{\frac{d}{k} \times d}$, which are responsible for the entries of the rows in $\Theta_g$. 
We perform an element-wise multiplication of these vectors with our $h_g'$, \ie{} $\Gamma = h_g' \odot \Psi$, where $\Gamma \in \mathbb{R}^{d \times \frac{d}{k}}$. A shared non-linear MLP network is then used to expand each column of $\Gamma$ to the dimension $d$. This process yields the linear weight $\Theta_g \in \mathbb{R}^{d \times d}$. Consequently, $\Theta_g$ is determined solely based on the input transformation. 
\cred{This design fundamentally differs from the hypernetwork proposed in SIE~\cite{garrido2023sie}, which independently predicts all $d^2$ elements of the transformation matrix.  In contrast, COPE uses the shared learnable embedding vectors to generate the columns of $\Theta_g$, which significantly reduces the number of trainable parameters. Furthermore, the proposed conditional pseudo-negatives generation explicitly penalizes invariant collapse, allowing us to avoid the architectural constraint of the bias-free hypernetwork required by SIE~\cite{garrido2023sie}.}

We generate pseudo-negatives using \transnet{} by sampling a set of random rotations $\{{g_r}\}_{r=1}^M$. We uniformly sample quaternions representing 3D rotations to ensure coverage over the rotation space. Specifically, for each pseudo-negative, we generate a random vector $\mathbf{v}_r = (w_r, x_r, y_r, z_r)$ by sampling from a standard normal distribution $ \mathbf{v}_r \sim \mathcal{N}(0, I_4)$. We then normalize $\mathbf{v}_r$ to obtain a unit quaternion $g_r$ and ensure the scalar part is nonnegative: $ g_r = \frac{\mathbf{v}_r}{|\mathbf{v}_r|} \cdot \text{sign}(w_r).$ This approach ensures that the set of quaternions ${g_r}$ is uniformly distributed over the space of 3D rotations. Random rotations are fed to the \transnet{} for generating pseudo-negatives as in Eq. \ref{eq: gen_pseudo_neg}.
\\
\subsection{Transformer Backbone}
We employ a shared Siamese encoder with a point cloud transformer backbone $f$, initialized with weights from an MAE in which there are two branches with shared weights. \coolname{} operates on pairs of point clouds: a randomly sampled point cloud $\mathbf{x}_i$ and its transformed point cloud $\mathbf{x}_i^{+}$ with the applied transformation $g$. Following the standard transformer architecture on the point cloud, we first extract a sequence of non-overlapping point patches, which are then converted to a set of tokens. 

Given an input point cloud $\mathbf{x}$, we first used Furthest Point Sampling (FPS) to sample a set of $n$ center points. Then, K Nearest-Neighbour (KNN) is utilized to select $k$ neighboring points around each center point to construct $n$ point patches. Following this, patch tokens are generated using a mini-PointNet~\cite{pang2022masked} and fed to the transformer encoder. However, in \coolname{}, not all the patch tokens are processed by the encoder. A random masking strategy is applied independently to $\mathbf{x}_i$ and $\mathbf{x}_i^{+}$, masking $m$ number of the point patches. Only visible tokens are fed to an encoder ($\mathcal{E}$). This independent masking strategy encourages the learning of representations that are robust to noise and occlusions. Then, the encoded embeddings of visible tokens from the two inputs are sent to the decoders with shared weights. An extra learnable [CLS] token is appended to each encoder output. We implement a lightweight decoder $\mathcal{D}$. The objective of the decoder is to aggregate the encoder outputs into representations $\mathbf{z}_i$ and $\mathbf{z}_i^{+}$.

\begin{algorithm}[t!]
\KwData{Learning rate \( \epsilon = 0.01\)}
\KwData{Source and target embeddings \( \mathbf{z}_{src}, \mathbf{z}_{tgt} \)}
Initialize \( \mathcal{L}_{min} = \infty \)  %

\For{j in \(N_{out}\)}{
Initialize \( g_r = (w_r, x_r, y_r, z_r)\)\\
  \For{k in \(N_{in}\)}{
    \( \mathbf{z}^{g_r} = \mathrm{\transnet{}}(g_r) \mathbf{z}_{src}\) \\
    \( \mathbf{z}^{g_{r}^{-1}} = \mathrm{\transnet{}}(g_{r}^{-1})\mathbf{z}_{tgt} \)\\
    \( \mathcal{L}_{g_r} =  \left\|\mathbf{z}_{tgt} - \mathbf{z}^{g_r}\right\|_2^2 +
    \left\|\mathbf{z}_{src} - \mathbf{z}^{g_r^{-1}}\right\|_2^2 \)\\
    \( g_{r}^* = g_{r} - \epsilon \nabla_{g_r} L_{g_r} \)  \\
    \( g_r = \frac{g_r}{\|g_r\|} \cdot \text{sign}(w_r^*) \) \label{alg:normalization}
  }
  
  \If{\(\mathcal{L}_{g_r} < \mathcal{L}_{min}\)}{
    Update \( \mathcal{L}_{min} = \mathcal{L}_{g_r} \), set \( g_{est} = g_r \) \\
  }
}
\KwRet{$g_{\mathrm{est}}$}\
\caption{Optimization of Transformation Estimation with Minimum Loss Tracking.}
\label{algo}
\end{algorithm}

\subsection{Relative Pose Estimation Using \transnet{}}
To tackle the challenge of relative pose estimation between source and target point clouds, we introduce a novel algorithm that utilizes the \transnet{} network to optimize the transformation parameters directly. The algorithm focuses on estimating the quaternions to minimize the difference between the transformed source and target embeddings, thus achieving precise alignment.

The process, detailed in Algorithm \ref{algo}, starts by exploring a set of randomly generated quaternions $\{g_r\}$, representing potential rotations between the source and target. Each quaternion is evaluated in a bi-directional manner, applying the transformation to both the source and the target embeddings via $g_r$ and $g_r^{-1}$. Here $g_r^{-1}$ is the inverse of $g_r$. The quaternions are refined through an iterative process, where each is adjusted according to the gradient update weighted by $\epsilon=0.01$. After updating the quaternions, it is essential to ensure they remain valid unit quaternions, which is achieved by the normalization step at line \ref{alg:normalization}. The quaternion associated with the minimum loss is designated as $g_{est}$ serving as the final solution for the alignment. This process ensures that the iterative refinement systematically approaches the global minimum of the loss landscape. In summary, this algorithm extends the capabilities of \transnet{} to solve relative pose estimation by optimizing quaternions to align global embeddings of point clouds.  
\cred{Although Algorithm \ref{algo} uses an iterative approach, the computational cost remains minimal because the optimization operates directly on pre-computed embeddings. It uses the lightweight COPE network to predict the displacements, avoiding repetitive backbone inference. Furthermore, the random initializations are independent and fully parallelizable on GPUs. This design ensures computational feasibility while avoiding the large rotational errors often occurring at high initial rotations.}

\begin{table}[!t]
    \centering
    \caption{Shape classification results on ModelNet40 dataset and ScanObjNN's \texttt{OBJ-BG} subset under z/z and SO(3)/SO(3) evaluation scenario.}
    \begin{tabular}{lccc}
    \toprule
    \multirow{2}{*}{Method} & \multicolumn{2}{c}{ModelNet40} & ScanObjNN--\texttt{OBJ-BG} \\ \cmidrule(r){2-3} \cmidrule(r){4-4}
    & z / z & SO(3) / SO(3) & SO(3) / SO(3)\\ \midrule
    \multicolumn{4}{l}{Supervised} \\ \midrule
    PointNet \cite{qi2017pointnet} & 89.2 & 75.5 & 54.7\\
    PointNet++ \cite{qi2017pointnet} & 89.3 & 85.0 & 47.4\\
    PCT \cite{guo2021pct} & 90.3 & 88.5 & 45.8 \\     %
    SFCNN \cite{rao2019spherical} & 91.4 & 90.1 & --\\
    RIConv \cite{zhang2019rotation} & 86.5 & 86.4 & 78.1\\
    RI-GCN \cite{kim2020rotation} & 89.5 & 89.5 & 80.6\\
    GCAConv \cite{zhang2020global} & 89.0 & 89.2 & 80.3\\
    RI-Framework \cite{li2021rotation} & 89.4 & 89.3 & 79.9\\
    VN-DGCNN \cite{deng2018ppfnet} & 89.5 & 90.2 & 80.3\\
    OrientedMP \cite{luo2022equivariant} & 88.4 & 88.9 & 77.2\\
    Yu et al. \cite{yu2023rethinking}& 91.0 & 91.0 & 86.3 \\ 
    PaRot \cite{zhang2023parot}  & 90.9 & 90.8 & 82.6 \\
    LocoTrans \cite{chen2024local} & 91.6 & \textbf{91.5} & 84.5 \\ \midrule
    \multicolumn{4}{l}{Self-Supervised} \\ \midrule
    PointBERT \cite{yu2022point} & 91.5 & 89.6 & 83.8 \\
    PointMAE \cite{pang2022masked}& 90.7 & 89.3 & 83.3\\ 
    PointM2AE \cite{zhang2022point} & 91.7 & 90.2 & 84.5 \\
    Point2Vec \cite{abouzeid2023point2vec} & 91.3 & 88.4 & 84.7 \\
    PointGPT-S \cite{chen2023pointgpt} & 90.6 & 88.3 & 80.2 \\
      \rowcolor{green!15}\textbf{\coolname{}} & \textbf{92.5} & 91.1 & \textbf{89.0}\\
      \bottomrule
    \end{tabular}
    \label{tab: ours vs others modelnet scannet}
\end{table}

\section{Results}
In this section, we present results on point cloud benchmarks and a complementary 3D Invariant-Equivariant Benchmark image benchmark. We begin with shape classification on ModelNet40 and ScanObjectNN and relative pose estimation at the object and scene level. 
For a fair comparison with equivariant losses reported only in the image domain, we test our proposed loss and \transnet{} with a ResNet\mbox{-}18 backbone and evaluate on 3D Invariant-Equivariant Benchmark (3DIEBench) under identical training and evaluation protocols.

\subsection{Experimental Setup}

\begin{table}[t!]
\centering
\caption{Pretraining hyperparameters for \coolname{}.}
\label{tab:pretrain_hparams}
\begin{tabular}{@{}lc@{}}
\toprule
\textbf{Config} & \textbf{Value} \\
\midrule
Optimizer & AdamW \\
Weight decay & 0.05 \\
Learning rate & $1 \times 10^{-3}$ \\
LR schedule & Cosine annealing with linear warmup \\
Warmup epochs & 80 \\
Batch size & 512 \\
Training epochs & 1600 \\
$\beta$  & 0.3 \\
Encoder depth & 12 \\
Num. pseudo-negatives & 8 \\
\bottomrule
\end{tabular}
\end{table}
We initialize the encoder $\mathcal{E}$ of our network \( f \) using a point cloud MAE \cite{abouzeid2023point2vec} and pretrain with our contrastive formulation. The proposed \coolname{} is a modular component that integrates with any point cloud MAE backbone.  
We follow the procedure utilized in Masked Autoencoders for images~\cite{jiang2023layer, haghighat2024ropim} as an effective way of integrating the contrastive objective with mask reconstruction pre-training. 
We use the training set of the ShapeNet dataset \cite{chang2015shapenet}, which contains approximately 42,000 synthetic point clouds across 55 object categories for training. We sample 1024 points from each point cloud for pretraining. 

We evaluate our model on downstream tasks, specifically shape classification and relative pose estimation for both objects and scene levels, to demonstrate the advantage of the invariant and equivariant properties of our approach. Details of our hyperparameters and pre-training settings are given in Tab. \ref{tab:pretrain_hparams}. 

\subsection{Shape Classification}
We evaluate our model on two prominent 3D object classification datasets, namely, ModelNet40 and ScanObjectNN. ModelNet40 \cite{wu20153d} consists of 12,311 clean 3D CAD models from 40 categories, split into 9,843 instances for training and 2,468 for testing.
ScanObjectNN \cite{uy2019revisiting} is a more challenging dataset derived from real-world indoor scenes containing approximately 15,000 objects across 15 categories. 

\label{sec: shape classification}

We follow the protocols of \cite{esteves2018learning} for the evaluation in Tab. \ref{tab: ours vs others modelnet scannet}: z/z -- training and testing under rotations around the z-axis and SO(3)/SO(3) --  training and testing the under the arbitrary 3D rotations. 
Tab. \ref{tab: ours vs others modelnet scannet} provides a detailed comparison of different point cloud backbones applied to the ModelNet40 and ScanObjectNN datasets across different protocols. The table is divided into sections detailing Supervised and SSL approaches. Here supervised section consists of equivariant methods \cite{rao2019spherical} - \cite{chen2024local} which are designed to handle rotations. 
Our SSL-based model, \coolname{}, achieves superior performance on ModelNet40 and ScanObjNN, outperforming all supervised and self-supervised methods under SO(3)/SO(3) on challenging ScanObjNN \textit{OBJ-BG}  by a considerable margin, and ModelNet40 under z/z while remaining highly competitive under SO(3)/SO(3).
This demonstrates that our novel pretraining approach enhances the generalization capabilities across diverse ranges.

\subsection{Relative Pose Estimation}

\label{sec: pose estimation}
\begin{table*}[!ht]
    \caption{Relative pose estimation results over five categories with E$^2$PN protocol. Here, all the other compared methods are supervised. Among them, KPConv is a non-equivariant method while EPN and E$^2$PN are equivariant networks. Our self-supervised approach performs better than the supervised equivariant methods.}
    \label{tab: pose estim}
    \centering
    \resizebox{1\linewidth}{!}{%
    \setlength{\tabcolsep}{0.4em}
    \begin{tabular}{@{}llccccc@{}}
    \toprule
     \textbf{Learning Type} & \textbf{Mean / Max / Med (\textdegree)} & \multicolumn{1}{c}{Airplane} & \multicolumn{1}{c}{Bottle} & \multicolumn{1}{c}{Car} & \multicolumn{1}{c}{Chair} & \multicolumn{1}{c}{Sofa} \\ \midrule
    \multirow{3}{*}{Supervised Learning} & KPConv \cite{thomas2019kpconv} & 12.0 / 70.5 / 10.0 & 8.0 / 104.4 / 5.0 & 35.8 / 175.6 / 20.9 & 26.0 / 168.3 / 16.3 & 84.2 / 177.3 / 75.2 \\
     & EPN \cite{chen2021equivariant}  &  1.3 / 6.2 / 1.1 & \textbf{1.2} / 22.3 / \textbf{0.8} & 2.6 / 117.6 / 1.1 & \textbf{1.2} / 9.0 / \textbf{1.0} & 1.5 / 15.0 / 1.1 \\
     & E$^2$PN \cite{zhu2023e2pn}  &  1.5 / 11.1 / 1.1 & 1.9 / 46.1 / 1.1 & 3.3 / 74.0 / 1.5 & 2.9 / 38.2 / 1.8 & 2.7 / 33.6 / 1.9 \\ 
     \cdashline{1-7}
     \addlinespace[0.8ex]
    \rowcolor{green!15} Self-Supervised Learning & \textbf{\coolname{}} &  \textbf{0.8} / \textbf{2.2} / \textbf{0.8} & 1.7 / \textbf{9.6} / \textbf{0.8} & \textbf{1.4} / \textbf{3.7} / \textbf{1.3} & 1.5 / \textbf{5.8} / 1.4 & \textbf{1.2} / \textbf{4.0} / \textbf{1.1} \\
    \bottomrule
    \end{tabular}}
\end{table*}
We assess our model on relative pose estimation benchmark tasks, where the goal is to predict the relative rotation between pairs of point clouds.
We fine-tune the network $f$ and \transnet{} using $\mathcal{L}_{\text{align}^*}$ and $\mathcal{L}_{\text{cope}}$, and conduct inference following Algorithm \ref{algo}. The evaluation is conducted under two scenarios: object-level and scene-level point clouds. We use the mean
isotropic rotation error~\cite{yew2020rpm} for the evaluation.

\subsubsection{Object Level}
\begin{figure*}[!t]
    \centering
    \begin{minipage}[t]{0.48\linewidth}
        \centering
        \includegraphics[width=\linewidth]{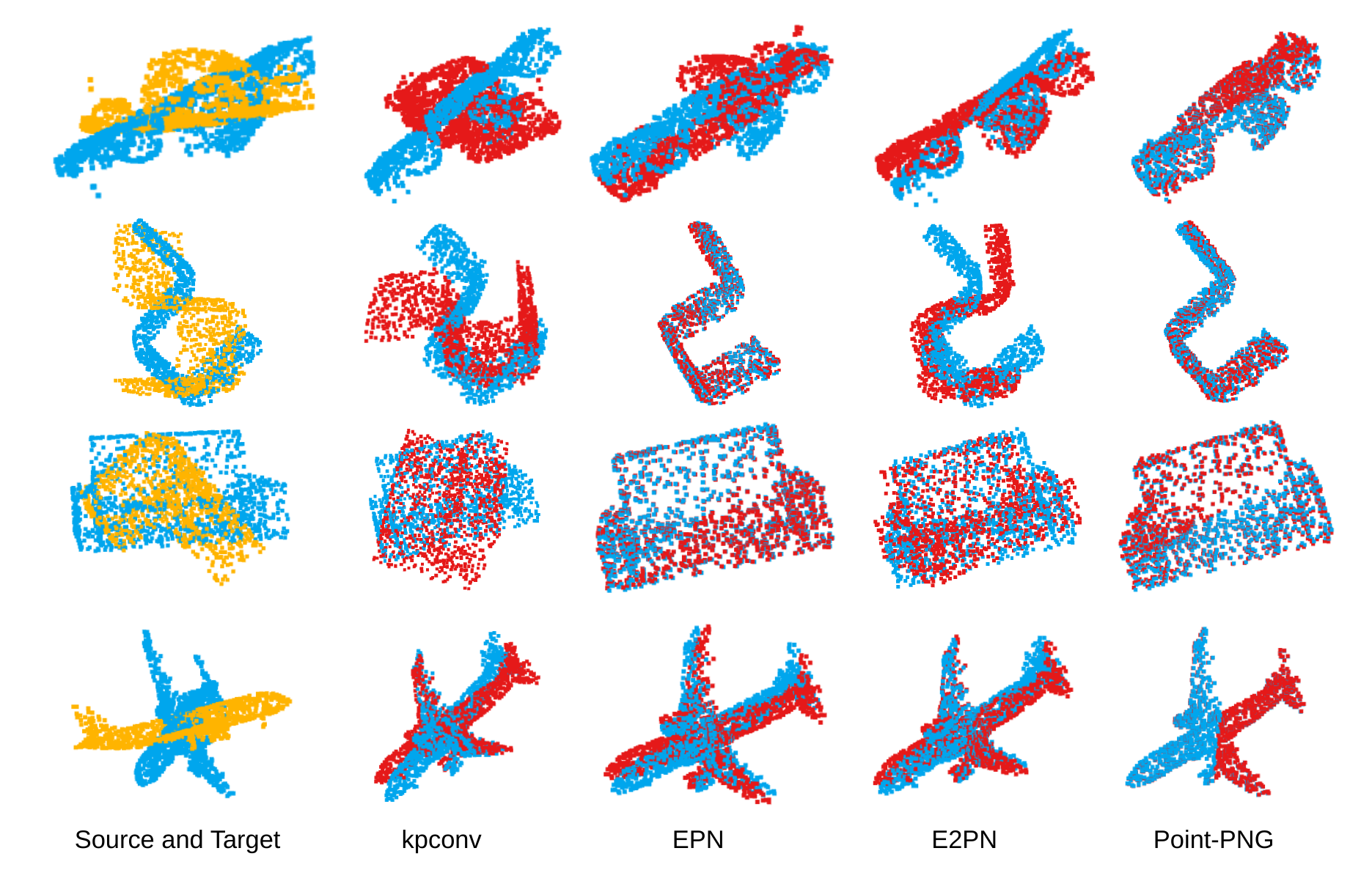}
        \caption{Illustration of relative pose estimation between source (yellow) and target (blue) point clouds for KPConv, EPN, E$^2$PN, and \coolname{}. Red indicates transformed source clouds.}
        \label{fig:modelnet-pose}
    \end{minipage}
    \hfill
    \begin{minipage}[t]{0.48\linewidth}
        \centering
        \includegraphics[width=\linewidth, trim=0cm 3cm 0cm 2.8cm, clip]{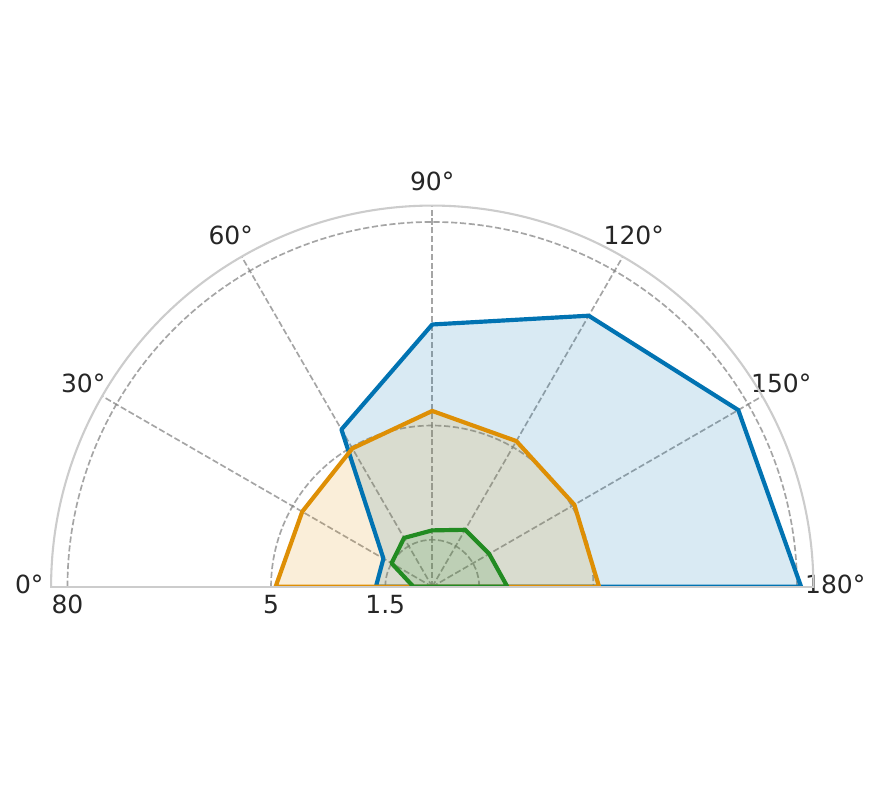}
        \caption{\cred{Relative pose estimation on 7Scenes evaluated using the mean rotation error under varying maximum rotational angles. The radial axis displays the error values on a logarithmic scale for better visualization. The angular axis represents the specified maximum rotational angle, where we generate random rotations by uniformly sampling the angles within the specified maximum. The plot compares\textcolor{cyan}{FMR~\cite{Huang_2020_CVPR}}, \textcolor{orange}{EquivReg~\cite{zhu2022correspondence}}, \textcolor{forestgreen}{\coolname{}}. (Lower is better)}}
        \label{fig:reg-7scene}
    \end{minipage}
\end{figure*}

In the object-level evaluation, we utilize five categories from the ModelNet40 dataset, following the protocol established by EPN~\cite{chen2021equivariant}. We present both quantitative and qualitative evaluations in Tab. \ref{tab: pose estim} and Fig. \ref{fig:modelnet-pose}, respectively.
Our self-supervised model, \coolname{}, shows the best overall performance compared to supervised counterparts, including both non-equivariant (such as KPConv~\cite{thomas2019kpconv}) and equivariant (EPN~\cite{chen2021equivariant} and E$^2$PN~ \cite{zhu2023e2pn}) networks. By achieving lower maximum rotation errors across all categories, \coolname{} demonstrates its superior ability to learn robust rotation-sensitive embeddings. This helps to significantly enhance the alignment accuracy of point clouds, validating the effectiveness of our proposed method in handling all ranges of rotations.

\subsubsection{Scene Level}
In addition to the object-level evaluation, we extend our analysis to relative pose estimation at the scene level using the 7Scenes dataset, a real-world indoor dataset that is a subset of 3DMatch~\cite{zeng20173dmatch}.
This dataset features over 100,000 points per scene. We sample 1024 points to form the source and target point clouds. We compare our method with correspondence-free pose estimation methods where FMR~\cite{Huang_2020_CVPR} is an iterative approach and EquivReg~\cite{zhu2022correspondence} uses an equivariant encoder. Following the protocol described EquivReg~\cite{zhu2022correspondence}, we generate random rotations by uniformly sampling the angles with the specified maximum $\left( 0^\circ,\ldots,180^\circ\right)$, alongside randomly selected rotation axes, to compute rotation errors.
The findings are summarized in Fig. \ref{fig:reg-7scene} with a logarithmic scale for better visualization. It shows the \coolname{}'s ability to maintain consistent lower error across the range of angles.

\subsection{Ablation Study} 
\begin{table}[!t]
\caption{Ablation on number of pseudo‐negatives. \cred{Predictor Accuracy (PA) measures how accurately COPE estimates the localized displacement caused by the transformation, while Absolute Equivariance (AE) verifies the model sensitivity to transformations. ($\uparrow$) indicates higher is better}}
\label{tab:ablation_pseudo}
\centering
\begin{tabular}{cccc}
\toprule
$M$ & PA (↑)   & AE (↑)   & SO3/SO3 (↑) \\
\midrule
4  & 0.92 & 0.41 & 0.8955 \\
\rowcolor{green!15}
8  & 0.97 & 0.67 & 0.8991 \\
16 & 0.92 & 0.64 & 0.8951 \\
32 & 0.86 & 0.66 & 0.8999 \\
\bottomrule
\end{tabular}
\end{table}

\begin{table}[!t]
\caption{Ablation on the $\beta$ hyperparameter. \cred{Predictor Accuracy (PA) measures how accurately COPE estimates the localized displacement caused by the transformation, while Absolute Equivariance (AE) verifies the model sensitivity to transformations. ($\uparrow$) indicates higher is better}}
\label{tab:ablation_beta}
\centering
\begin{tabular}{cccc}
\toprule
$\beta$ & PA ($\uparrow$)   & AE ($\uparrow$)   & SO3/SO3 ($\uparrow$) \\
\midrule
0.9 & 0.81 & 0.79 & 0.8971 \\
0.7 & 0.82 & 0.70 & 0.9011 \\
0.5 & 0.89 & 0.64 & 0.8991 \\
\rowcolor{green!15}
0.3 & 0.97 & 0.67 & 0.8991 \\
0.0 & 0.97 & 0.00 & 0.8963 \\
\bottomrule
\end{tabular}
\end{table}

Following EquiMod [3], we use Absolute Equivariance (AE) to verify model sensitivity to transformations instead of collapse to invariance. When AE equals zero, $\Theta_g$ acts as the identity and the representation becomes strictly invariant. We also measure Predictor Accuracy (PA) as how accurately COPE estimates the localized displacement caused by the transformation. Their definitions are

\begin{equation}
\label{eq:ae_pa}
\begin{split}
    AE ={}& \operatorname{sim}\left(\frac{\Theta_g\,\mathbf{z}_i}{\|\Theta_g\,\mathbf{z}_i\|},\,\mathbf{z}_i^{+}\right)  - \operatorname{sim} \left(\mathbf{z}_i,\,\mathbf{z}_i^{+}\right); \\
    PA ={}& \operatorname{sim} \left(\frac{\Theta_g\,\mathbf{z}_i}{\|\Theta_g\,\mathbf{z}_i\|},\,\mathbf{z}_i^{+}\right),
\end{split}
\end{equation}
where $\operatorname{sim}(\cdot,\!\cdot)$ denotes cosine similarity. Relative pose estimation between point clouds according to Algorithm \ref{algo} relies on iteratively aligning global embeddings using the displacement predicted by COPE. High PA ensures the learned displacement faithfully models the true latent shift while high AE guarantees that distinct rotations yield sufficiently distinct embeddings.

Tables \ref{tab:ablation_pseudo} and \ref{tab:ablation_beta} show that choosing eight pseudo‐negatives ($M=8$) together with a loss weight of $\beta=0.3$ achieves the highest Predictor Accuracy (PA = 0.97) and Absolute Equivariance (AE = 0.67). When we freeze the encoder and probe with an MLP, classification accuracy varies by no more than 0.004 across all settings, demonstrating that explicitly separating negatives for the uniformity loss $\mathcal{L}_{\text{uniform}}$ and the COPE loss $\mathcal{L}_{\text{cope}}$ preserves discriminative power. In contrast, omitting $\mathcal{L}_{\text{cope}}$ (i.e.\ setting $\beta=0$) drives AE to zero, confirming that our proposed loss is essential to prevent collapse into a fully invariant solution. 

\subsection{Qualitative analysis}
\begin{figure}[htbp]
    \centering
    \begin{minipage}[c]{0.75\linewidth}
        \centering
        \includegraphics[width=\linewidth]{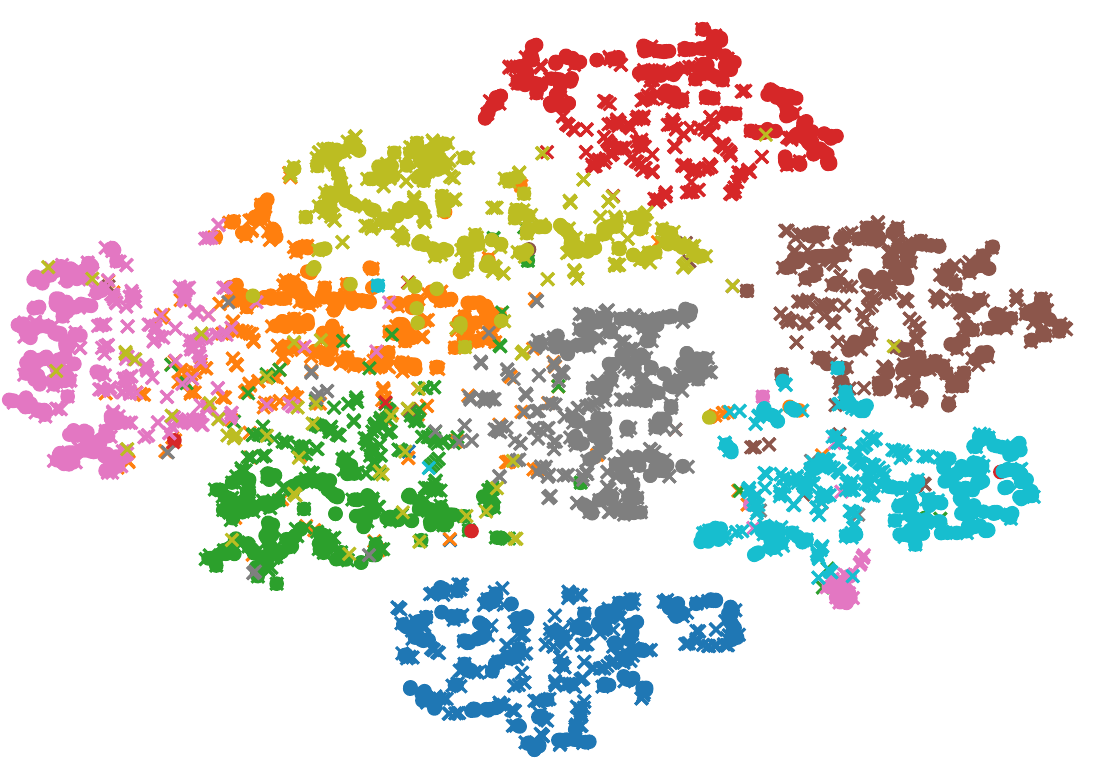}
    \end{minipage}
    \hfill   
    \begin{minipage}[c]{0.23\linewidth}
        \centering
        \includegraphics[width=\linewidth]{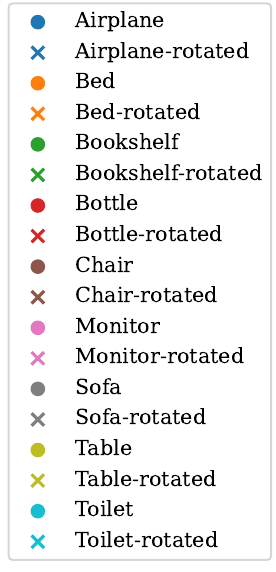}
    \end{minipage}
    \caption{t-SNE Visualisation of learned features from ModelNet40 dataset by our proposed method \coolname{}. The circular markers indicate features of non-rotated point cloud and the `cross' markers indicate features of rotated point cloud. Every class is visualised with a unique colour.}
 \label{fig: tsne_all_class}
\end{figure}

We conduct t-SNE visualisation of features learned by our proposed method on rotated and non-rotated point clouds. Fig. \ref{fig: tsne_all_class} shows the t-SNE graph. In the graph, we show the features of selective classes from the ModelNet40 dataset. We compare the features of point clouds when they are rotated with when they are in their canonical pose (\ie{} not rotated). We apply random rotations to the point clouds. The graph clearly shows that our learned features of rotated point clouds across different classes shift from their canonical counterparts. Even after the shift, they still form non-noisy and meaningful clusters that resonate with the underlying class. 

\begin{figure}[t]
    \centering
    \includegraphics[width=0.49\linewidth]{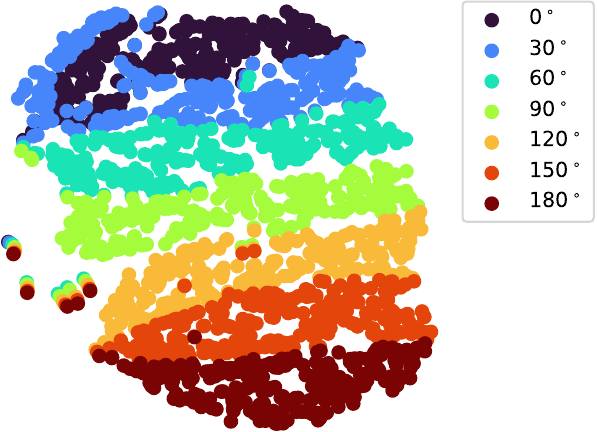}
        \caption{t-SNE Visualisation of rotated point cloud features by our proposed method \coolname{}. Each point cloud under the chair category is rotated at an interval of $30^\circ$ by six times.}
    \label{fig: tsne_single_class}
\end{figure}

For a further understanding, we test our method using a large pool of point cloud rotations under the chair category. Fig. \ref{fig: tsne_single_class} shows the t-SNE graph. We select the point cloud samples from the chair category and compare six rotations at different angles and no rotation. It is clear to see that every rotation shifts our features by a margin. Visually, the larger the rotation, the larger the shift will be. However, the shifted features maintain the class structure as indicated by the chair category point clouds in Fig. \ref{fig: tsne_all_class}. This suggests that our proposed method is able to effectively learn both discriminative and equivariance features at the same time. 

\subsection{Comparisons with SOTA Equivariant Representation Learning Methods in 3DIEBench}

In this experiment, we rigorously evaluate our proposed $\mathcal{L}_{\coolname}$ and its predictor module \transnet{} against existing state-of-the-art equivariant representation learning methods in the image domain.
These methods are not designed for direct use on point clouds
Specifically, we compare our approach with SEN~\cite{park2022sen}, EquiMod~\cite{devillers2023equimod}, and SIE~\cite{garrido2023sie} on the established 3DIEBench~\cite{garrido2023sie}.
As \coolname{} is originally developed for point clouds, we adapt it to the image domain and ensure a fair evaluation by using the same ResNet-18 backbone architecture across all methods. 

3DIEBench is specifically designed to evaluate both invariant and equivariant properties of learned representations. It contains over 2.5 million images generated by rendering 52,472 objects from ShapeNetCore~\cite{chang2015shapenet} across 55 categories under a diverse set of 3D rotations.
We follow the protocol established in SIE~\cite{garrido2023sie} and use the provided image splits without modification. This setup allows consistent evaluation across tasks such as classification and rotation prediction, where the objective is to assess how well representations capture class-specific discriminative features and transformation sensitivity.

\begin{table}[t!]
    \centering
    \caption{Quantitative evaluation of learned representations on invariant (classification) and equivariant (rotation prediction) tasks. For fair comparison, we adopt \coolname{}$^{*}$ by adding our proposed loss and predictor module to a ResNet-18 backbone, following the same architecture used in prior works.}
    \begin{tabular}{@{}lcc@{}}
        \toprule
        Method & Classification (top-1) ($\uparrow$) & Rotation prediction ($R^2$) ($\uparrow$) \\
        \midrule
        SEN~\cite{park2022sen} & 86.93 & 0.51 \\
        EquiMod~\cite{devillers2023equimod} & 87.19 & 0.47 \\
        SIE~\cite{garrido2023sie} & 82.94 & 0.73 \\
        \coolname{}$^{*}$ & \textbf{87.7} & \textbf{0.74} \\
        \bottomrule
    \end{tabular}
    \label{tab:class}
\end{table}

\subsubsection{Classification and rotation prediction}
We follow the protocol of SIE in 3DIEBench dataset and assess representation quality on image classification and rotation prediction tasks. For classification, we train a linear classifier on frozen representations, and for rotation prediction, we regress the transformation between two views using a 3-layer MLP. As shown in Table~\ref{tab:class}, \coolname$^{*}$ outperforms prior methods, achieving the highest top-1 classification accuracy (87.70\%) and $R^2$ score for rotation prediction (0.74), indicating strong discriminative features and transformation sensitivity.\\

\begin{table}[t!]
\centering
\caption{Quantitative evaluation on MRR, H@1, and H@5 metrics.}
\begin{tabular}{lccc}
\toprule
 Method & MRR ($\uparrow$) & H@1 ($\uparrow$) & H@5 ($\uparrow$) \\
\midrule
EquiMod~\cite{devillers2023equimod} & 0.16 & 0.05 & 0.22 \\
SEN$^*$~\cite{park2022sen} & 0.17 & 0.05 & 0.22 \\
SIE~\cite{garrido2023sie} & 0.41 & 0.30 & 0.51 \\
\coolname{}$^{*}$ & \textbf{0.54} & \textbf{0.41} & \textbf{0.69} \\
\bottomrule
\end{tabular}
\label{tab:mrr_h1_h5}
\end{table}
\begin{figure*}[!ht]
\centering
\renewcommand{\arraystretch}{1.4}
\setlength{\tabcolsep}{4pt}

\begin{tabular}{c c c c c c c}
& \textbf{Source} & \textbf{Target} & \textbf{NN1} & \textbf{NN2} & \textbf{NN3} \\

\rotatebox{90}{\small\hspace{2.8em}SIE} &
\includegraphics[width=0.12\textwidth]{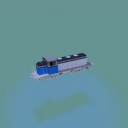} &
\includegraphics[width=0.12\textwidth]{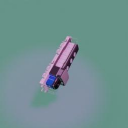} &
\includegraphics[width=0.12\textwidth]{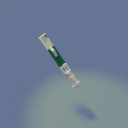} &
\includegraphics[width=0.12\textwidth]{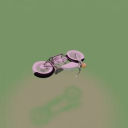} &
\includegraphics[width=0.12\textwidth]{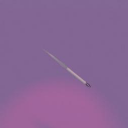} & \\

\rotatebox{90}{\small \coolname$^{*}$} &
\includegraphics[width=0.12\textwidth]{imgs/images/415408/source.pdf} &
\includegraphics[width=0.12\textwidth]{imgs/images/415408/target.pdf} &
\includegraphics[width=0.12\textwidth]{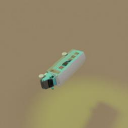} &
\includegraphics[width=0.12\textwidth]{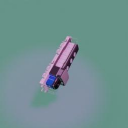} &
\includegraphics[width=0.12\textwidth]{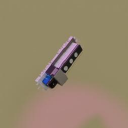} & \\
\\
\rotatebox{90}{\small\hspace{2.8em}SIE} &
\includegraphics[width=0.12\textwidth]{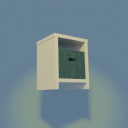} &
\includegraphics[width=0.12\textwidth]{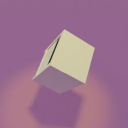} &
\includegraphics[width=0.12\textwidth]{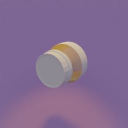} &
\includegraphics[width=0.12\textwidth]{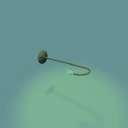} &
\includegraphics[width=0.12\textwidth]{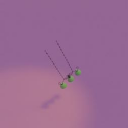} & \\

\rotatebox{90}{\small \coolname$^{*}$} &
\includegraphics[width=0.12\textwidth]{imgs/images/157614/source.pdf} &
\includegraphics[width=0.12\textwidth]{imgs/images/157614/target.pdf} &
\includegraphics[width=0.12\textwidth]{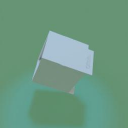} &
\includegraphics[width=0.12\textwidth]{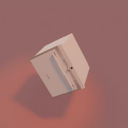} &
\includegraphics[width=0.12\textwidth]{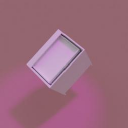} & \\
\\
\rotatebox{90}{\small\hspace{2.8em}SIE} &
\includegraphics[width=0.12\textwidth]{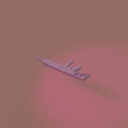} &
\includegraphics[width=0.12\textwidth]{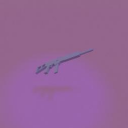} &
\includegraphics[width=0.12\textwidth]{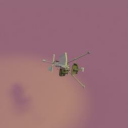} &
\includegraphics[width=0.12\textwidth]{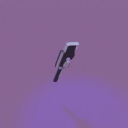} &
\includegraphics[width=0.12\textwidth]{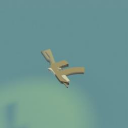} & \\

\rotatebox{90}{\small \coolname$^{*}$} &
\includegraphics[width=0.12\textwidth]{imgs/images/118383/source.pdf} &
\includegraphics[width=0.12\textwidth]{imgs/images/118383/target.pdf} &
\includegraphics[width=0.12\textwidth]{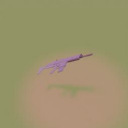} &
\includegraphics[width=0.12\textwidth]{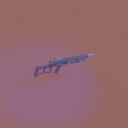} &
\includegraphics[width=0.12\textwidth]{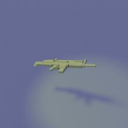} & \\
\\

\end{tabular}
\caption{ Nearest neighbours of predicted embeddings: Given the source embedding and the transformation to the target, each method’s predictor generates a predicted embedding whose top-3 nearest neighbours (NN1–NN3) are shown.}
\label{fig.retreiv}
\end{figure*}

\subsubsection{Predictor Evaluation via Image Retrieval}
To evaluate the effectiveness of our predictor, we adopt the retrieval-based metrics introduced in SIE~\cite{garrido2023sie}: Mean Reciprocal Rank (MRR), Hit Rate at 1 and 5 (H@1, H@5). Given a source embedding and the known transformation, we apply the learned predictor and retrieve the nearest neighbours among views of the same object. MRR measures the average inverse rank of the ground-truth target, while H@k reports if the target appears within the top-k neighbours. As reported in Table~\ref{tab:mrr_h1_h5}, \coolname{}$^{*}$ achieves superior retrieval performance (MRR = 0.54, H@1 = 0.41, H@5 = 0.69), significantly outperforming EquiMod and SIE. This confirms that our pseudo-negative formulation avoids predictor collapse and promotes a more transformation-aware embedding space. 
Notably, our predictor achieves these results using only 200K learnable parameters, which is significantly fewer than the 4 million parameters required by the SIE predictor. This efficiency is enabled by our use of a shared parameters across the columns of the predicted weight matrix in the COPE architecture. In contrast, SIE predicts all $d^2$ elements of the transformation matrix independently from the rotation parameters, leading to substantially higher model size.

\subsubsection{Qualitative evaluation on predictor accuracy}
To qualitatively evaluate the accuracy of the localized displacement caused by the predictor of our proposed method, \coolname{}$^{*}$, we conduct a nearest-neighbor retrieval analysis in the embedding space. Specifically, given a source object, the task involves predicting embeddings corresponding to the transformation of the target object using the predictor. Retrieval of the nearest neighbors of these predicted embeddings is performed across a search space comprising images from all classes, with the expectation that the retrieved neighbors closely match the target's pose and category.

Fig.~\ref{fig.retreiv} compares our method to SIE through representative examples. Results indicate that embeddings generated by SIE predictor frequently retrieve nearest neighbors that mismatch in either pose or category. In contrast, embeddings from our proposed \coolname{}$^{*}$ consistently exhibit superior alignment, closely matching both the pose and category of the target object.  These results highlight the improved performance of the \transnet{} predictor in modeling transformation-specific displacements and demonstrate the effectiveness of \coolname{} in learning transformation-sensitive representations.

\section{Conclusion}
\label{sec:conclusion}
In this paper, we proposed \coolname{}, a novel self-supervised learning framework aimed at preventing collapse to invariant solutions in contrastive learning for point cloud data. Our approach introduces a loss function that leverages pseudo-negatives, ensuring that the learned representations remain both discriminative and sensitive to transformations. We also presented a novel inference algorithm for relative pose estimation, which utilizes the transformation-sensitive feature space and \transnet{} network to iteratively estimate the relative pose between point clouds. We conducted extensive experiments on shape classification and relative pose estimation tasks, which demonstrate that \coolname{} outperforms existing methods, especially in challenging rotation scenarios. For relative pose estimation tasks, our framework excels in accurately estimating relative poses between point clouds, leading to superior performance compared to supervised baselines. 

\section{Limitation}
In this work, we primarily analyze 3D rotations as the transformation of interest. Extending \coolname{} to other transformation types such as scaling, translation, or deformation, and modeling their combined effects, remains an important direction for future work. Additionally, our approach assumes a linear relationship in the feature space caused by transformations, which may not fully capture complex or non-rigid transformation behaviors. Addressing these limitations could further improve the generality and expressiveness of our framework.

\appendix[Proof]
\section{\break Degenerate identity predictor is a global optimum for the SIE loss}
In this subsection, we demonstrate that the SIE loss~\cite{garrido2023sie} admits a trivial global optimum where the predictor collapses to the identity mapping and the encoder outputs invariant, non-collapsed features. Specifically, we show that this degenerate solution achieves minimum loss, thus failing to prevent the collapse to trivial invariance. In contrast, our proposed loss function, incorporating pseudo-negatives explicitly, penalizes such identity-predictor collapse. We prove that our loss attains the maximum (in contrast to the minimum), precisely under the degenerate identity solution, enforcing the model to learn non-trivial, transformation-sensitive representations.

To formalize this, consider the case in which:
\begin{enumerate}
    \item The predictor collapses to the identity mapping, and
    \item The encoder produces \emph{invariant, non–collapsed} features.
\end{enumerate}

\paragraph{SIE objective.}
Given an encoder \(f_\theta\!:\!\mathcal{X}\!\to\!\mathbb{R}^d\) and a
predictor \(\rho_\psi:\mathcal{G}\!\to\!\mathrm{GL}(d)\), the SIE loss
for a pair of views
\((x_i,x'_i)=(x_i,g_i\!\cdot\!x_i)\) decomposes into

\begin{equation}
\label{eq:loss:sie}
\begin{split}
\mathcal{L}_{\text{SIE}}
&=\underbrace{\mathcal{L}_{\text{reg}}(Z)\!+\!\mathcal{L}_{\text{reg}}(Z')}_{\text{regularization}}
\\ &\quad
 +\lambda_{\text{inv}}\,\underbrace{\frac{1}{N}\!\sum_{i=1}^N \lVert Z_{i,\text{inv}} Z'_{i,\text{inv}}\rVert_2^2}_{\mathcal{L}_{\text{inv}}}
\\ &\quad
  + \lambda_{\text{eq}}\,\underbrace{\frac{1}{N}\!\sum_{i=1}^N\lVert \rho_\psi(g_i)Z_{i,\text{eq}} -Z'_{i,\text{eq}}\rVert_2^2}_{\mathcal{L}_{\text{equiv}}}
\\ &\quad
  + \lambda_V\,\underbrace{V\!\bigl(\rho_\psi(g_i)Z_{i,\text{eq}}\bigr)}_{\mathcal{L}_{\text{stab}}}\,.
\end{split}
\end{equation}

where \(Z_i=f_\theta(x_i)=[Z_{i,\text{inv}},\,Z_{i,\text{eq}}]\).  
\begin{align*}
\mathcal{L}_{\text{reg}}(Z) &= \lambda_C \, C(Z) + \lambda_V \, V(Z), \quad \text{with} \\
C(Z) &= \frac{1}{d} \sum_{i \ne j} \text{Cov}(Z)^2_{i,j}, \quad \text{and} \\
V(Z) &= \frac{1}{d} \sum_{j=1}^{d} \max \left( 0, 1 - \sqrt{\text{Var}(Z_{\cdot,j})} \right).
\end{align*}

\vspace{0.5em}
\paragraph{Constructing the degenerate solution.}
Assume there exists an encoder \(f_{\theta^\star}(\cdot)\) that outputs
\emph{invariant} but non–collapsed features:
\[
\underbrace{f_{\theta^\star}(x_i)}_{Z_i}
 =\underbrace{f_{\theta^\star}(g_i\!\cdot\!x_i)}_{Z'_i}
 =:Z_i^\star
 \quad \forall\,x_i,\;g_i\in\mathcal{G}.
\tag{C1}
\label{cond:invariance}
\]
Because the features are non‐collapsed, each dimension has above unit variance and zero cross‐covariance, hence
\(C(Z^\star)=V(Z^\star)=0\).

Let the predictor collapse to the identity,
\[
\rho_{\psi^\star}(g)=I_d
\quad \forall\,g\in\mathcal{G}.
\tag{C2}
\label{cond:predictor_collapse}
\]

\begin{itemize}
\item \textbf{Regularization.}\;
      \(\mathcal{L}_{\text{reg}}(Z_i^\star)=0\) by construction.
\item \textbf{Invariant term.}\;
      \(Z_{i,\text{inv}}^\star=Z_{i,\text{inv}}^{\star\,\prime}\)
      implies \(\mathcal{L}_{\text{inv}}=0\).
\item \textbf{Equivariant term.}\;
      \(\rho_{\psi^\star}(g_i)=I_d\) gives \\
      \(\mathcal{L}_{\text{equiv}}
        =\frac{\lambda_{\text{eq}}}{N}\sum_i
         \lVert Z_{i,\text{eq}}^\star-Z_{i,\text{eq}}^\star\rVert_2^2=0\).
\item \textbf{Stability term.}\;
      Because \(V(Z_i^\star)=0\), \\ also
      \(V(\rho_{\psi^\star}(g_i)Z_{i,\text{eq}}^\star)=V(I_dZ_{i,\text{eq}}^\star)=0\).
\end{itemize}

Every component of~\eqref{eq:loss:sie} therefore vanishes, so
\(\mathcal{L}_{\text{SIE}}(\theta^\star,\psi^\star)=0\).  
Hence an identity predictor together with fully invariant features is a
(global) minimizer of the SIE objective

\vspace{0.75em}
\paragraph{Our Loss Penalizes Predictor Collapse}
Here we verify that the same degenerate configuration
\((\theta^\star,\psi^\star)\) is \emph{not} optimal for our  objective. Fig. \ref{fig: pseudo_negatives} conceptually illustrates our objective, where for each input embedding $\mathbf{z}_i$, we generate an anchor $\frac{\Theta_{g_i}\mathbf{z}_i}{\|\Theta_{g_i}\mathbf{z}_i\|}$ using its corresponding transformation $g_i$, along with multiple pseudo-negatives $\tilde{\mathbf{z}}_{i}^{g_r}$, each corresponding to a random transformation $g_r$. Here $\mathbf{z}_i$ and $\mathbf{z}_i^{+}$ are the embeddings of original and the transformed inputs, both normalized to unit norm.
Both the anchor and pseudo-negatives are generated by our predictor \transnet{}, parameterized by $\psi$. 

Our overall objective is defined as:
\begin{equation}
    \mathcal{L}_{\coolname} = \mathcal{L}_{\text{align}^*} +  \beta \mathcal{L}_{\text{cope}} + (1 - \beta) \mathcal{L}_{\text{unif}},
\end{equation}
\begin{equation}
    \mathcal{L}_{\text{align}^*} = \frac{1}{N} \sum_{i=1}^{N} \left\|\frac{\Theta_{g_i}\mathbf{z}_i}{\|\Theta_{g_i}\mathbf{z}_i\|} - \mathbf{z}_i^{+}\right\|_2^2,
\end{equation}
\begin{equation}
    \mathcal{L}_{\text{unif}} = \log \mathbb{E}_{(\mathbf{z}_i, \mathbf{z}_k)\sim \mathcal{Z}\times\mathcal{Z}} e^{-\left\|\mathbf{z}_i - \mathbf{z}_k\right\|_2^2 / \tau},
\end{equation}
where \(\Theta_{g_i} =\transnet{}_{\psi}(g_i)\),
\paragraph{Alignment and uniformity.}
Conditions~\eqref{cond:invariance},\eqref{cond:predictor_collapse}
drive the alignment loss to its minimum,  
\(\mathcal{L}_{\text{align}^\star}=0\), and likewise minimize
\(\mathcal{L}_{\text{unif}}\).

\paragraph{The COPE term resists collapse.}
\begin{equation}\label{eq:l_cope}
\begin{split}
\mathcal{L}_{\text{cope}}
&= \frac{1}{N} \sum_{i=1}^{N}\log \Biggl[\sum_{r=1}^{M} \exp\left(-\frac{\left\| \frac{\Theta_{g_i} \mathbf{z}_i}{\| \Theta_{g_i} \mathbf{z}_i \|} - \tilde{\mathbf{z}}_{i}^{g_r} \right\|_2^2}{\tau}\right)
\\
&\qquad\qquad\quad
      + \exp\left(-\frac{\left\| \frac{\Theta_{g_i} \mathbf{z}_i}{\| \Theta_{g_i} \mathbf{z}_i \|} - \mathbf{z}_i^{+} \right\|_2^2}{\tau}\right) \Biggr].
\end{split}
\end{equation}
Here $\tilde{\mathbf{z}}_{i}^{g_r}=\frac{\Theta_{g_r} \mathbf{z}_i}{\| \Theta_{g_r} \mathbf{z}_i \|}$ are pseudo negatives corresponding to random transformation $g_r$.
\begin{equation} 
\label{eq: l_cope} 
\begin{split}
\mathcal{L}_{\text{cope}} &= \frac{1}{N} \sum_{i=1}^{N}\log \Biggl[\sum_{r=1}^{M} \exp\left(-\frac{\left\| \frac{\Theta_{g_i} \mathbf{z}_i}{\| \Theta_{g_i} \mathbf{z}_i \|} - \frac{\Theta_{g_r} \mathbf{z}_i}{\| \Theta_{g_r} \mathbf{z}_i \|} \right\|_2^2}{\tau}\right) 
\\
&\qquad\qquad\quad
+ \exp\left(-\frac{\left\| \frac{\Theta_{g_i} \mathbf{z}_i}{\| \Theta_{g_i} \mathbf{z}_i \|} - \mathbf{z}_i^{+} \right\|_2^2}{\tau}\right) \Biggr]
\end{split}
\end{equation}

We substitute \(\Theta_{g_i}=\Theta_{g_r} = I_d\) to simulate the collapse case: Here the predictor \transnet{} outputs the same identity weight for every transformation.
\begin{equation} 
\label{eq: l_cope} 
\mathcal{L}_{\text{cope}} = \frac{1}{N} \sum_{i=1}^{N}\log \left[M + \exp\left(-\frac{\left\| \frac{\mathbf{z}_i}{\| \mathbf{z}_i \|} - \mathbf{z}_i^{+} \right\|_2^2}{\tau}\right) \right]
\end{equation}
As  $\|\mathbf{z}_i\|=1$, we have:

\begin{equation} 
\label{eq: l_cope} 
\mathcal{L}_{\text{cope}} = \frac{1}{N} \sum_{i=1}^{N}\log \left[M + \exp\left(-\frac{\left\| \mathbf{z}_i - \mathbf{z}_i^{+} \right\|_2^2}{\tau}\right) \right]
\end{equation}
Under the Condition~\eqref{cond:invariance}, where the encoder is invariant, $\mathbf{z}_i$ and $\mathbf{z}_i^{+}$ are equal. Therefore, we have:
\[
\mathcal{L}_{\text{cope}} = \frac{1}{N} \sum_{i=1}^{N}\log\Bigg[\sum_{r=1}^{M} 1 + 1\Bigg] = \log(M+1),
\]
For any predictor outputs satisfying 
\[
\Theta_{g_r}\neq\Theta_{g}\quad\text{for all }r=1,\dots,M,
\]
\begin{equation}
\begin{split}
    \left\| \frac{\Theta_{g_i} \mathbf{z}_i}{\| \Theta_{g_i} \mathbf{z}_i \|} - \frac{\Theta_{g_r} \mathbf{z}_i}{\| \Theta_{g_r} \mathbf{z}_i \|} \right\|_2^2 &> 0 
    \\[0.15em]
    -\frac{\left\| \frac{\Theta_{g_i} \mathbf{z}_i}{\| \Theta_{g_i} \mathbf{z}_i \|} - \frac{\Theta_{g_r} \mathbf{z}_i}{\| \Theta_{g_r} \mathbf{z}_i \|} \right\|_2^2}{{\tau}} &< 0
\end{split}
\end{equation}
\begin{equation}
\begin{split}
    \exp\left(-\frac{\left\| \frac{\Theta_{g_i} \mathbf{z}_i}{\| \Theta_{g_i} \mathbf{z}_i \|} - \frac{\Theta_{g_r} \mathbf{z}_i}{\| \Theta_{g_r} \mathbf{z}_i \|} \right\|_2^2}{\tau}\right) &< 1
    \\[0.15em]
    \sum_{r=1}^{M} \exp\left(-\frac{\left\| \frac{\Theta_{g_i} \mathbf{z}_i}{\| \Theta_{g_i} \mathbf{z}_i \|} - \frac{\Theta_{g_r} \mathbf{z}_i}{\| \Theta_{g_r} \mathbf{z}_i \|} \right\|_2^2}{\tau}\right) &< M
\end{split}
\end{equation}

\begin{equation}
\mathcal{L}_{\text{cope}} < \log(M+1).
\end{equation}
$\mathcal{L}_{COPE}$ becomes its maximum when the predictor collapses to identity.
Hence our loss explicitly penalizes the identity‐predictor degeneration, forcing the optimizer to learn a non‐trivial, transformation sensitive mapping.

\vspace{0.5em}

The above analysis shows that SIE alone cannot prevent the identity solution, whereas the COPE term in $\mathcal{L}_{\coolname}$ introduces a penalty that impose the learning dynamics towards equivariant representations.

\section*{Acknowledgment}
This work was supported partly by an Australian Research Council Discovery Project Grant DP250103634 and the CSIRO Embodied AI Research Cluster. 
\balance{}

\bibliographystyle{IEEEtran}
\bibliography{ref}
\begin{IEEEbiography}[{\includegraphics[width=1in,height=1.25in,clip, keepaspectratio]{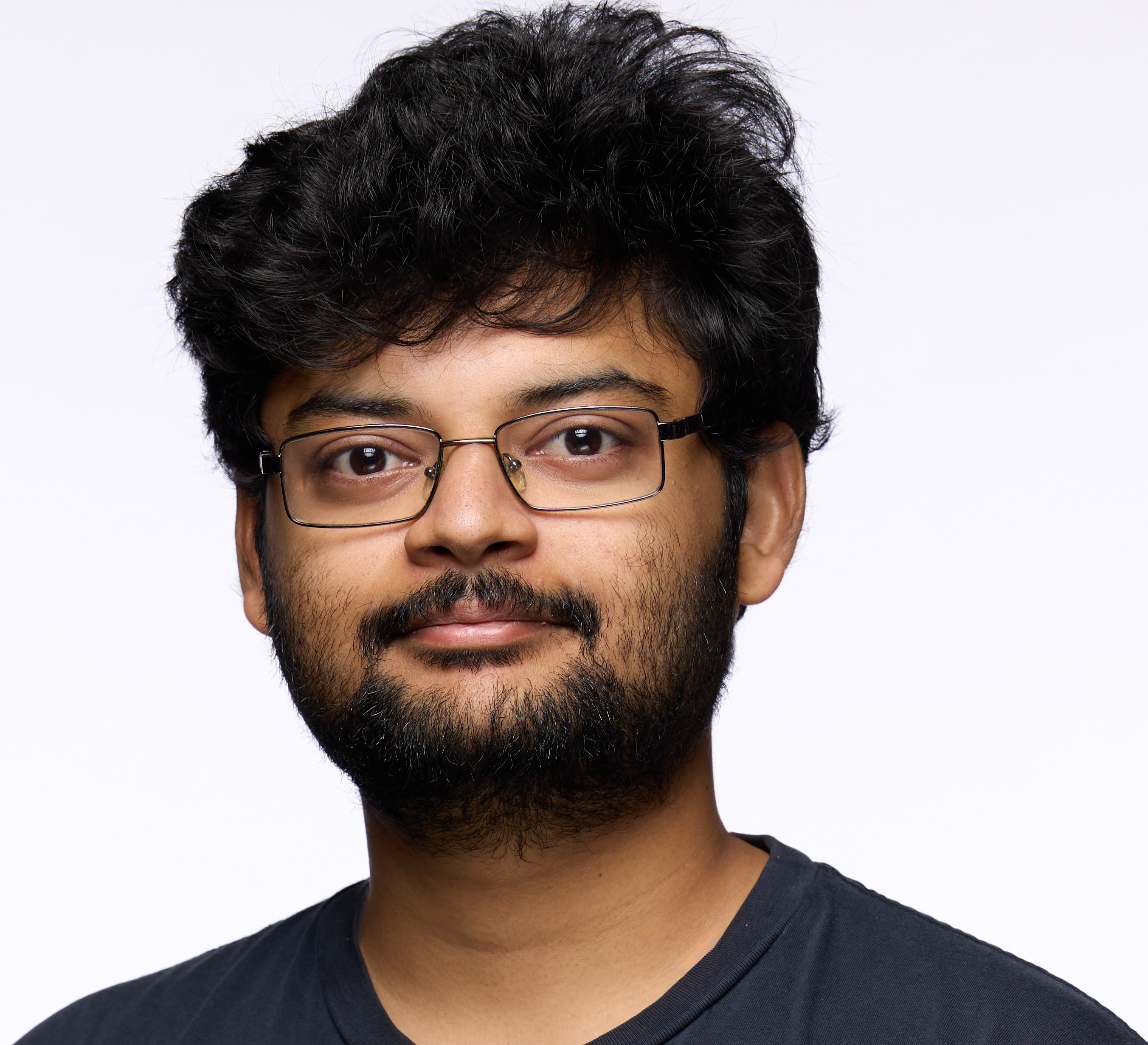}}]{Sutharsan Mahendren}
(Student Member, IEEE) received the B.Sc. degree in Electronic and Telecommunication Engineering from the University of Moratuwa, Sri Lanka, in 2020. He is currently pursuing the Ph.D. degree with the Queensland University of Technology (QUT), Brisbane, QLD, Australia, and CSIRO, DATA61, Brisbane. His research interests include computer vision and self-supervised learning for 3D perception.
\end{IEEEbiography}

\begin{IEEEbiography}[{\includegraphics[width=1in,height=1.25in,clip, keepaspectratio]{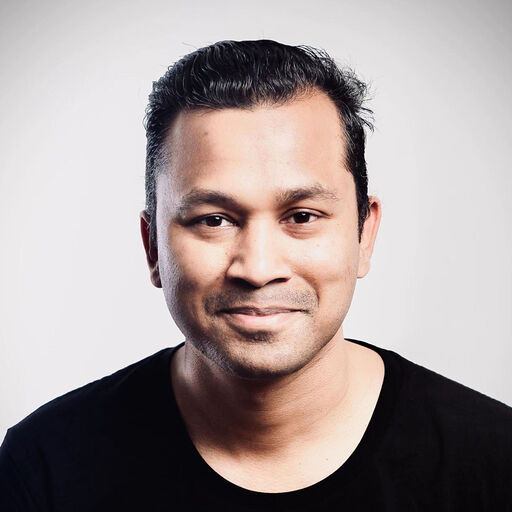}}]{SAIMUNUR RAHMAN} is a Research Scientist in the Robotics Group at CSIRO’s Data61. Previously, he was a postdoctoral researcher in the Embodied AI Cluster at CSIRO. He received his M.Sc. (by Research) degree in 2017 from Multimedia University, Malaysia, and his Ph.D. in 2023 from the Visual Informatics for Learning and Applications (VILA) group at the University of Wollongong, Australia, both with a focus on visual representation learning. His research interests include visual navigation, kernel methods, and deep learning. He has also served on program committees of international conferences in computer vision and machine learning.

\end{IEEEbiography}

\begin{IEEEbiography}[{\includegraphics[width=1in,height=1.25in,clip,keepaspectratio]{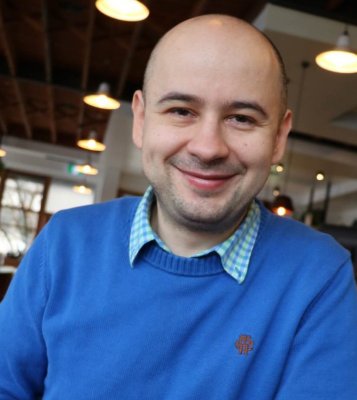}}]{PIOTR KONIUSZ}
received the BSc degree in Telecommunications and Software Engineering from Warsaw University of Technology, Poland, in 2004, and the PhD degree in Computer Vision from CVSSP, University of Surrey, U.K., in 2013. He is a Senior Researcher with the Machine Learning Research Group, Data61/CSIRO, and a Senior Honorary Lecturer at the Australian National University (ANU). He was previously a postdoctoral researcher with the LEAR team at INRIA, France. His research interests include representation learning (contrastive and self-supervised learning), VLMs, MMLMs, LLMs, and deep and graph
neural networks, as well as image classification and action recognition. He has received awards including the Sang Uk Lee Best Student Paper Award (ACCV 2022), Runner-up APRS/IAPR Best Student Paper Award (DICTA 2022). He served as a Workshop Program Chair for NeurIPS 2023 and The WebConf 2025, and is a Program Chair for NeurIPS 2025.
\end{IEEEbiography}

\begin{IEEEbiography}[{\includegraphics[width=1in,height=1.25in,clip,keepaspectratio]{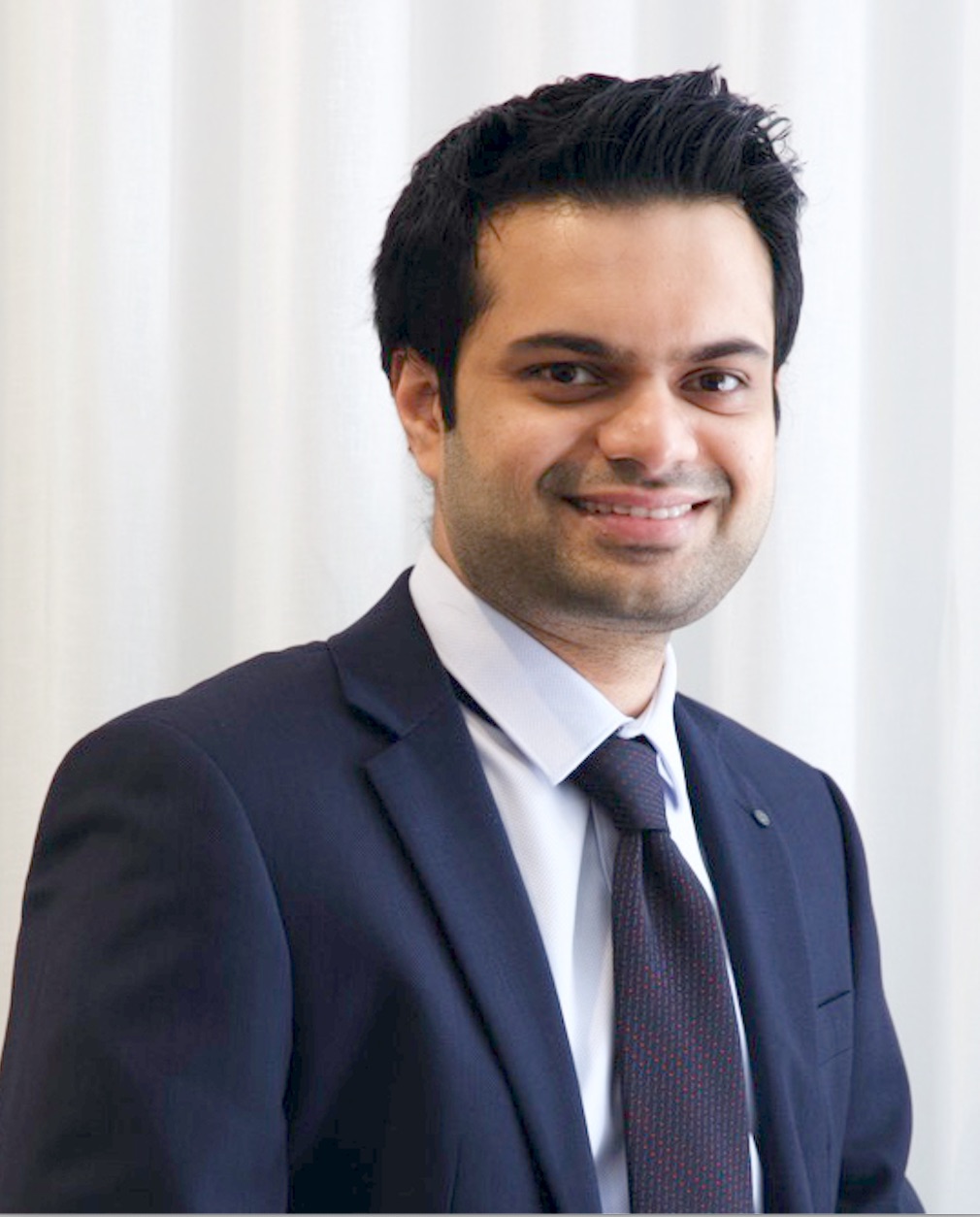}}]{Tharindu Fernando}
(Member, IEEE)  received his BSc (special degree in computer science) from the University of
Peradeniya, Sri Lanka, and his PhD from Queensland University of Technology (QUT), Australia.
He is currently a Research Fellow in the Signal Processing, Artificial Intelligence, and
Vision Technologies (SAIVT) research program at the School of Electrical Engineering and Robotics
at Queensland University of Technology (QUT). He is a recipient of the 2019 QUT University Award for
Outstanding Doctoral Thesis, the QUT Early Career Researcher Award in 2022, the QUT Faculty of Engineering Early Career Achievement Award in 2024, and the 2024 National Intelligence Post-Doctoral Grant. His research interests include Artificial Intelligence, Computer Vision, Deep Learning, Bio Signal Processing, and Video Analytics.
\end{IEEEbiography}

\begin{IEEEbiography}[{\includegraphics[width=1in,height=1.25in,clip,keepaspectratio]{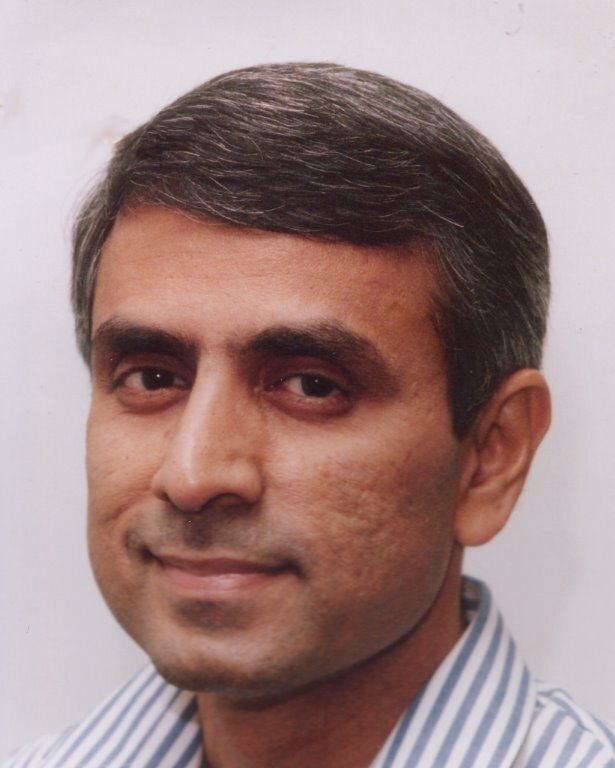}}]{Sridha Sridharan }
(Life Senior Member, IEEE) (Life Senior Member, IEEE) received his MSc degree in
communication engineering from the University of Manchester, UK, and his PhD
degree in the area of signal processing from the University of New South Wales,
Australia. He is currently with the Queensland University of Technology (QUT),
where he is a Professor in the School of Electrical Engineering and Robotics. He is
also the Co-Director of the Research Program in Signal Processing, Artificial
Intelligence and Vision Technologies (SAIVT) at QUT. He has published over 300
papers consisting of publications in journals and in refereed international
conferences in the areas of signal processing, computer vision and machine
learning, and has graduated 85 PhD students at QUT in these areas. He has also
received a number of research grants from various funding bodies including
Commonwealth competitive funding schemes, such as the Australian Research
Council (ARC), Cooperative Research Centres (CRC) and the National Security
Science and Technology (NSST) unit. Several of his research outcomes have been
commercialised.
\end{IEEEbiography}

\begin{IEEEbiography}[{\includegraphics[width=1in,height=1.25in,clip,keepaspectratio]{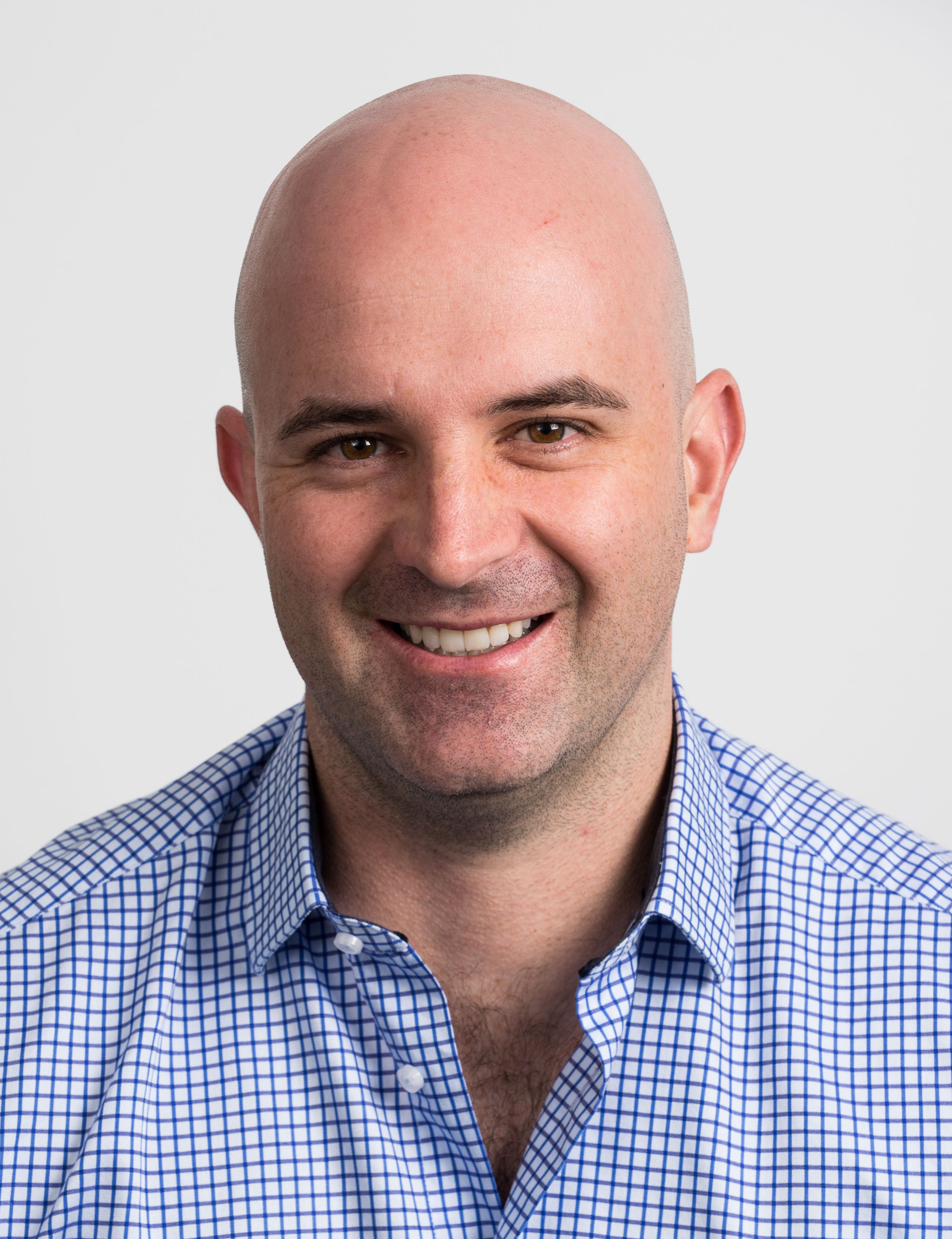}}]{Clinton Fookes}
(Senior Member, IEEE) received the B.Eng., M.B.A., and Ph.D. degrees in aerospace/ avionics from the Queensland University of Technology (QUT), Brisbane, QLD, Australia, in 1999, 2011, and 2004, respectively. He is a Professor in Vision \& Signal Processing, the Associate Dean of Research of the Faculty of Engineering at QUT, and is the Director of Signal Processing, Artificial Intelligence and Vision Technologies (SAIVT). He is a Fellow of the Australian Academy of Technological Sciences and Engineering, Fellow of the International Association of Pattern Recognition, and Fellow of the Asia-Pacific Artificial Intelligence Association. He serves on the editorial boards for the IEEE Transactions on Image Processing, and Pattern Recognition, and has previously served on the editorial board for the IEEE Transactions on Information Forensics \& Security. Prof. Fookes is a multi-award winning researcher including an Australian Institute of Policy and Science Young Tall Poppy, an Australian Museum Eureka Prize, Engineers Australia Engineering Excellence Award, Australian Defence Scientist of the Year, and a Senior Fulbright Scholar.
\end{IEEEbiography}

\begin{IEEEbiography}[{\includegraphics[width=1in,height=1.25in,clip,keepaspectratio]{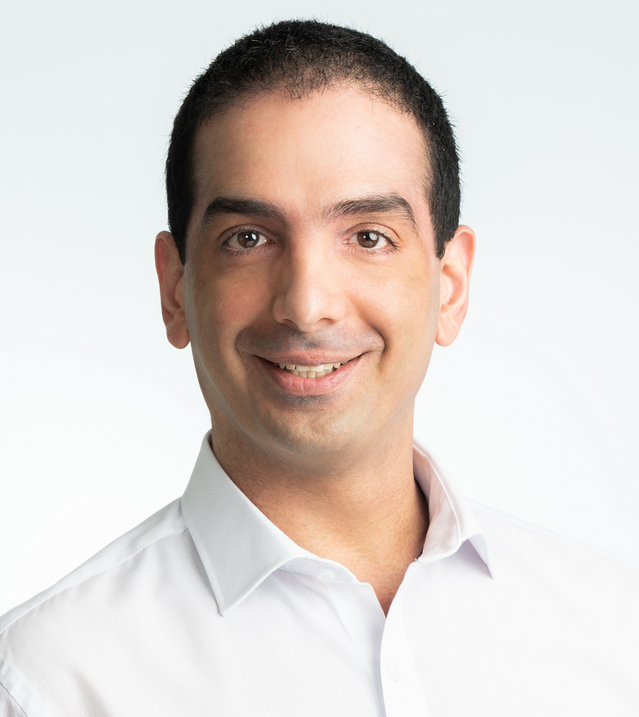}}]{Peyman Moghadam} (Senior Member, IEEE) is a Senior Principal Research Scientist with CSIRO, DATA61, Brisbane, QLD, Australia, and a Professor (Adjunct) with the Queensland University of Technology (QUT), Brisbane. Currently, he is the head of the Embodied AI Research Cluster at CSIRO, working at the intersection of Robotics and Machine learning. From 2020-2024, he led the Spatiotemporal AI portfolio at CSIRO’s Machine Learning and Artificial Intelligence (MLAI) Future Science Platform, advancing MLAI methods for scientific discovery in spatiotemporal data streams. In 2022, he was a Visiting Professor at ETH Zürich. Peyman is a Senior Member of IEEE, a Member of ACM, and an ACM Distinguished Speaker. His research interests include embodied AI, robotics, and self-supervised learning.
\end{IEEEbiography}

\EOD

\end{document}